\documentclass[sigconf]{acmart}
\acmConference[KDD '24]{30th ACM SIGKDD Conference on Knowledge Discovery and Data Mining}{August 25--August 29, 2024}{Barcelona, Spain}

\usepackage{subcaption}
\usepackage{multirow}
\usepackage{enumitem}
\usepackage{tabularx}
\usepackage[utf8]{inputenc}
\usepackage{amsmath}
\usepackage{amsthm}
\usepackage{microtype}
\usepackage[english]{babel}
\usepackage{wasysym}
\usepackage{graphicx}
\usepackage[linesnumbered,ruled,vlined]{algorithm2e}
\SetKwInput{KwInput}{Input}                
\SetKwInput{KwOutput}{Output}  
\newcommand{\annotator}[2]{\csdef{#1}##1{{\color{#2} [\textbf{\MakeUppercase #1}: ##1]}}}
\annotator{sunxin}{red}

\AtBeginDocument{%
  \providecommand\BibTeX{{%
    \normalfont B\kern-0.5em{\scshape i\kern-0.25em b}\kern-0.8em\TeX}}}

\copyrightyear{2024} \acmYear{2024} \setcopyright{acmlicensed}\acmConference[KDD '24]{Proceedings of the 30th ACM SIGKDD Conference on Knowledge Discovery and Data Mining}{August 25--29, 2024}{Barcelona, Spain}\acmBooktitle{Proceedings of the 30th ACM SIGKDD Conference on Knowledge Discovery and Data Mining (KDD '24), August 25--29, 2024, Barcelona, Spain}\acmDOI{10.1145/3637528.3671878}\acmISBN{979-8-4007-0490-1/24/08}

\begin{document}

\title[DIVE]{DIVE: Subgraph Disagreement for Graph Out-of-Distribution Generalization}

\author{Xin Sun}
\orcid{0000-0003-4459-4245}
\affiliation{%
  \institution{University of Science and Technology of China}
  \city{Hefei}
  \country{China} \\
    \institution{NLPR, MAIS,\\Institute of Automation, Chinese Academy of Sciences}
  \city{Beijing}
  \country{China}}
\email{sunxin000@mail.ustc.edu.cn}

\author{Liang Wang}
\affiliation{%
  \institution{NLPR, MAIS,\\Institute of Automation, Chinese Academy of Sciences}
  \city{Beijing}
  \country{China}}
\email{liang.wang@cripac.ia.ac.cn}

\author{Qiang Liu}
\authornote{To whom correspondence should be addressed.}
\affiliation{
  \institution{NLPR, MAIS,\\Institute of Automation, Chinese Academy of Sciences}
  \city{Beijing}
  \country{China}
  }
\email{qiang.liu@nlpr.ia.ac.cn}

\author{Shu Wu}
\affiliation{%
  \institution{NLPR, MAIS,\\Institute of Automation, Chinese Academy of Sciences}
  \city{Beijing}
  \country{China}}
\email{shu.wu@nlpr.ia.ac.cn}

\author{Zilei Wang}
\affiliation{%
  \institution{University of Science and Technology of China}
  \city{Hefei}
  \country{China}}
\email{zlwang@ustc.edu.cn}

\author{Liang Wang}
\affiliation{%
  \institution{NLPR, MAIS,\\Institute of Automation, Chinese Academy of Sciences}
  \city{Beijing}
  \country{China}}
  \email{wangliang@nlpr.ia.ac.cn}

\renewcommand{\shortauthors}{Xin Sun et al.}
\begin{abstract}
This paper addresses the challenge of out-of-distribution (OOD) generalization in graph machine learning, a field rapidly advancing yet grappling with the discrepancy between source and target data distributions. Traditional graph learning algorithms, based on the assumption of uniform distribution between training and test data, falter in real-world scenarios where this assumption fails, resulting in suboptimal performance. A principal factor contributing to this suboptimal performance is the inherent simplicity bias of neural networks trained through Stochastic Gradient Descent (SGD),  which prefer simpler features over more complex yet equally or more predictive ones. This bias leads to a reliance on spurious correlations, adversely affecting OOD performance in various tasks such as image recognition, natural language understanding, and graph classification. Current methodologies, including subgraph-mixup and information bottleneck approaches, have achieved partial success but struggle to overcome simplicity bias, often reinforcing spurious correlations. To tackle this, our study introduces a new learning paradigm for graph OOD issue. We propose DIVE, training a collection of models to focus on all label-predictive subgraphs by encouraging the models to foster divergence on the subgraph mask, which circumvents the limitation of a model solely focusing on the subgraph corresponding to simple structural patterns. Specifically, we  employs a regularizer to punish overlap in extracted subgraphs across models, thereby encouraging different models to concentrate on distinct structural patterns. Model selection for robust OOD performance is achieved through validation accuracy. Tested across four datasets from  GOOD benchmark and one dataset from DrugOOD benchmark, our approach demonstrates significant improvement over existing methods, effectively addressing the simplicity bias and enhancing generalization in graph machine learning. 
\end{abstract}



\maketitle

\section{Introduction}
The rapid advancement of graph machine learning has opened up a myriad of opportunities and challenges, particularly in the realm of distribution shift between source and target data.  Most existing graph learning algorithms work under the statistical assumption that the training and test data are drawn from the same distribution. However, this assumption does not hold in a lot of real-word scenarios, where the source data fails to adequately represent the target domain's characteristics, leading to suboptimal performance and generalization issues. 

\textbf{Simplicity bias degrades generalization ability.} Neural networks trained using Stochastic Gradient Descent (SGD) have been recently demonstrated to exhibit a preference for simple features, while neglecting equally predictive or even more predictive complex features~\cite{Shah2020ThePO}. \textbf{This simplicity bias hinders the learning of complex patterns that constitute the core mechanisms of the task of interest. When these simple patterns are merely spurious correlations\cite{simon1954spurious, zhang2023mining}, the model's out-of-distribution (OOD) performance significantly deteriorates.} For instance, in image recognition, a typical example of a spurious correlation is the reliance on the background instead of the object's shape. In natural language understanding, it manifests as the preference for specific words rather than grasping the sentence's overarching meaning.  In graph-based tasks, a notable example is the focus on a molecule's scaffold rather than the functional groups that are actually important. Due to the mechanism of message passing, structural patterns with higher degrees and higher modularity are likely to receive more attention~\cite{DBLP:conf/aaai/LiuN023, DBLP:conf/www/ShomerJ0T23, DBLP:conf/cikm/TangYSWTAMW20}, resulting in the scaffold subgraph being simpler to learn. More specifically, in the context of predicting water solubility, the presence of cyclic structures is not the actual determinant of solubility. Although molecules with cyclic structures typically exhibit poorer water solubility, the actual determinants of solubility are the polar functional groups that confer polarity to the molecule, such as hydroxyl and amino groups. However, for a model trained using stochastic gradient descent (SGD), focusing on cyclic structures within a graph is typically simpler than concentrating on polar functional group structures. This can lead the model to overemphasize simple and spurious features while neglecting the learning of complex but causal features, significantly affecting the model's generalization ability.

\textbf{The pitfall of current methods under simplicity bias.} To address the issue of failure in out-of-distribution generalization, the most effective and widely adopted strategy at present is based on subgraph-mixup~\cite{wu2022discovering, liu2022graph, Jia2023GraphIL, Fan2022DebiasingGN, Xiang2023LearningIM}. Subgraph-mixup approach involves initially utilizing a subgraph extractor to identify the underlying invariant or causal subgraph, which maintains a consistent correlation with the target labels across various graph distributions from different environments. Subsequently, it combines the invariant subgraph with the spurious subgraph (the complement of invariant subgraphs) from another instance to augment the dataset and achieve improved results. \textbf{Although these methods have achieved some empirical success, the faithfulness of the extracted invariant subgraph is questionable in the presence of  simplicity bias.} Specifically, when the spurious subgraph is the simpler pattern and is equally predictive of training labels, the subgraph extractor faces challenges in extracting the invariant subgraph due to the simplicity bias. If the estimated invariant subgraph parts include spurious information, assigning the label corresponding to the invariant part to the mixuped graph is likely to reinforce the spurious correlation between the spurious subgraph and the labels.  Another line of work is based on information bottleneck (IB)~\cite{miao2022interpretable, Yu2020GraphIB}, achieving generalization by maximizing mutual information
between labels and invariant subgraphs while minimizing mutual
information between the subgraph and the entire graph. \textbf{However, the IB method does not inherently distinguish between causal relevance and spurious correlation between the extract subgraph and label~\cite{Hua2022Causal}.} This limitation can lead the IB method to retain spurious information of the extracted subgraph.  When the spurious subgraph is simpler pattern and equally predictive on the training set, this issue will become more severe.

\textbf{Our method}. Due to the presence of the simplicity bias in the training procedure, the current method are unable to accurately identify the correct subgraphs. Besides, given the inherent characteristics of the SGD training method, the simplicity bias is difficult to avoid. To address this issue, we propose \textbf{DIVE}, training a collection of models to attend to all predictive graphs with \textbf{DIVE}rsity regularization, which allows us to identify not only subgraphs with simple structural  patterns but also those with complex patterns. Specifically,  We propose training a collection of models of same architecture to fit the training data by focusing on different, yet label-predictive subgraphs. These label-predictive subgraphs encompass both spurious and invariant subgraphs. Each model undergoes optimization for standard empirical risk minimization (ERM), complemented by the use of a regularizer designed to penalize the overlap of extracted subgraphs across the collection. This strategy encourages each model to attend on diverse structural patterns within the graph data, rather than solely on the simplest ones. The process of identifying a model with robust OOD performance is reduced to an independent model selection step, for which we employ validation accuracy as the metric for model selection. Our method, tested across four datasets from the GOOD benchmark and one dataset from the DrugOOD benchmark, demonstrates significant improvement over existing approaches. 
Our main contributions can be summarized as follows:
\begin{itemize}[leftmargin=*]
    \item  We propose a novel paradigm to  address the out-of-distribution issue in graph tasks by learning a collection of diverse predictors, which is robust to the simplicity bias.
    \item  We introduce diversity among models in the collection through a novel subgraph mask diversity loss that encourage different models attend to different predictive subgraph. And our method is capable of extracting the invariant subgraph more precisely than the current methods because of the subgraph diversity regularization.
    \item We conduct comprehensive experiments on 5 datasets and the experimental results demonstrate the superiority of our method compared to the state-of-the-art approaches.
\end{itemize}
\section{Related Work}
\subsection{Diversity on Ensemble Models}
The diversity on ensemble models has been extensively explored on visual task to solve the distribution shift problem. While the diversity stage similarly learns a collection for diverse models, our approach differs in that we directly optimize for diversity on subgraph. The bias-variance-covariance decomposition~\cite{Ueda1996GeneralizationEO}, which generalizes the bias variance decomposition to ensembles, shows how the error decreases with the covariances of the member of the ensemble. Despite its importance, there is still no well accepted definition and understanding of diversity, and it is often derived from prediction errors of members of the ensemble. This creates a conflict between trying to increase accuracy of individual predictors $h$, and trying to increase diversity. In this view, creating a good ensemble is seen as striking a good balance between individual performance and diversity. To promote diversity in ensembles, a classic approach is to add stochasticty into training by using different subsets of the training data for each predictor~\cite{Breiman2004BaggingP}, or using different data augmentation methods~\cite{Stickland2020DiverseEI, Jain2023DARTDT}. Another approach is to add orthogonality constrains on the predictor's gradient~\cite{Ross2019EnsemblesOL, Kariyappa2019ImprovingAR, Teney2021EvadingTS}. the information bottleneck~\cite{Tishby2000TheIB} also has been used to promote ensemble diversity~\cite{Ram2021DICEDI, Sinha2020DIBSDI}. Recently, Some work claims that diversity can be achieved by  producing different prediction on out-of-distribution dataset~\cite{Pagliardini2022AgreeTD, Lee2023DiversifyAD}. 

However, the diversity on ensembles has rarely explored on graph out-of-distribution task. To this end, we propose to diversify a collection of models by allowing them to make different predictions on subgraph masks. Unlike the aforementioned methods, our approach, DIVE, can be trained on the full dataset and does not require any out-of-distribution data during training. Additionally, it does not impose constraints on the predictions of the classifier  but instead fosters model diversity through disagreement on subgraph mask predictions. Furthermore, in contrast to many previous models, our individual predictors do not share the same encoder, enhancing the diversity and robustness of our approach. 

It is noteworthy to mention that the paper is not about building ensembles. Ensembling means that the results from the diversity models are aggregated for inference. Rather, we train a collection of models and select on model for inference. The goal of ensembling is to combine models with uncorrelated errors into one of lower variance. Our goal is to discover all predictive patterns normally missed by the SGD learning because of the simplicity bias.

\subsection{Graph Out-of-Distribution Generalization}
Graph structure is ubiquitous in real world, such as molecular\cite{wigh2022review}, protein\cite{zhang2022protein}, social networks\cite{zhang2021mining} and knowledge graph\cite{xia2024enhancing, xia2023metatkg, zhang2023learning}. Graph representation learning\cite{chen2020graph, wang2024rethinking, zhu2021graph} achieves deep learning on graphs by encoding them into vector in a latent space. Despite their significant success, current Graph Neural Networks (GNNs) largely depend on the identically distributed (I.D.) assumption, meaning that the training and test data are drawn from the same distribution. However, in reality, various forms of distribution shifts often occur between training and testing datasets due to unpredictable data generation mechanisms, leading to out-of-distribution (OOD) scenarios.

Our research focuses on graph classification, where methods for out-of-distribution generalization are primarily classified into three categories. The foremost and extensively investigated strategy hinges on the concept of subgraph-mixup~\cite{wu2022discovering, liu2022graph, Jia2023GraphIL, Fan2022DebiasingGN, Xiang2023LearningIM}. The second category revolves around the principle of the information bottleneck~\cite{chen2022learning, Chen2023DoesIG, miao2022interpretable, Yu2020GraphIB}, achieving generalization by maximizing mutual information between labels and invariant subgraphs while minimizing mutual information between the subgraph and the entire graph.  But as mentioned before, these two categories of methods fail to extract the correct invariant subgraph in the presence of simplicity bias. The last category is invariant learning~\cite{yang2022learning, li2022learning, Yuan2023EnvironmentAwareDG}. These methods aim to find a invariant subgraph whose predictive relationship with the target values remains stable across different environments. However, these methods need environment labels  which is often unavailable and expensive to obtain on graphs. Some methods propose to infer the environment labels~\cite{yang2022learning, li2022learning}. However, the reliability of these estimated labels is pivotal. If the environment label estimate induce a higher bias or noise, it would make the learning of graph invariant patterns even harder~\cite{Chen2023DoesIG}.

\subsection{Simplicity Bias}
Deep learning is rigorously investigated to decipher the reasons behind its notable successes and occasional failures. Key concepts such as the simplicity bias, gradient starvation, and the learning of functions of increasing complexity have been instrumental in shedding light on the inherent lack of robustness in deep neural networks. These insights explain why performance can significantly degrade under minor distribution shifts and adversarial perturbations. Shah et al.~\cite{Shah2020ThePO} revealed that neural networks trained with Stochastic Gradient Descent (SGD) exhibit a tendency to prefer learning the simplest predictive features within the data, often at the expense of more complex, yet more predictive ones. Alarmingly, methods believed to enhance generalization and robustness, such as ensembles and adversarial training, have been shown to be ineffective in counteracting the simplicity bias. Recently, a lot of diversity-based ensembles methods~\cite{Pagliardini2022AgreeTD, Lee2023DiversifyAD, Teney2021EvadingTS, Jain2023DARTDT} are proposed to solve the the simplicity bias and gain empirical success.

\section{Method}
\subsection{Notions}
Denote an attributed graph as $G=(A, X)$, where $ A = \{0, 1\}^{n\times n}$ is the adjacent matrix and $X$ includes node attributes. $A_{ij} =1$ represents that there exists an edge between node $i$ and $j$, and $A_{ij}=0$ otherwise. The node set and the edge set can be denoted as $V$ and $E$, respectively. We focus on graph-level out-of-distribution  task and a dataset set of graphs can be denoted as $\{(G_i, Y_i)\}_{i=1}^N$, where $N$ is the number of samples in the training set. 
\subsection{Problem Formulation}
We consider a supervised learning setting in which we train a model $f$ that takes input $G \in \mathcal{G}$ and predicts its corresponding label $Y \in \mathcal{Y}$, where $\mathcal{G}$ and $\mathcal{Y}$ are graph space and label space respectively. Generally, we are given a set of datasets collected from multiple environments and each dataset $D^e$ contains pairs of input graph and its label: $D_e = \{G_i,Y_i\}_{i=1}^{N_e}$ drawn from the joint distribution $P_e(G, Y)$ of environment $e$. We define the training  dataset as $D_{train}$ that are drawn from the joint distribution $P_{train}(G, Y) = P_{e \in \mathcal{E}_{train} \subseteq \mathcal{E}_{all}}(G, Y)$, and the test dataset as $D_{test}$ that are drawn from the joint distribution $P_{test}(G, Y) = P_{e \in \mathcal{E}_{test}\subseteq \mathcal{E}_{all}}(G, Y)$.  We aims to find a  optimal predictor $f^*$ that minimize $\max_{e\in \mathcal{E}_{all}} R_e$, where $R_e$ is the empirical risk of $f$ under environment $e$ \cite{Vapnik1991PrinciplesOR, arjovsky2019invariant}. 

\subsection{Learning all label-predictive subgraphs}
We assume each graph instance $G_i$ consists of two parts of information, one part is invariant information, which is the determinants information of the task of interest. The other part is the spurious information. We assume the spurious information and the invariant information are all predictive to the labels of training set. While our method focus on the structural distribution shift, we only consider the structural label-predictive patterns. We assume that  a graph instance can have multiple spurious subgraphs and one invariant subgraph. The set of spurious subgraphs is represented as $\{G_i^{S_1}, G_i^{S_2}, \cdots, G_i^{S_n}\}$ and the invariant subgraph  is denoted as $G_i^I$. Consequently,  the set of all label-predictive subgraphs is symbolized as $S_{G_i} = \{G_i^{S_1}, G_i^{S_2}, \cdots, G_i^{S_n}, G_i^I\}$. To learn all these label-predictive subgraphs, thus circumventing simplicity bias,  we train a set $\mathcal{F}$ of near-optimal predictor $f: \mathcal{G} \rightarrow \mathcal{Y}$. Let $\mathcal{L}_p: \mathcal{F} \rightarrow \mathbb{R}$ be the risk with respect to $P_{train}(G, Y)$. The $\epsilon$-optimal set with respect to $\mathcal{F}$ as level $\epsilon \geq 0$ is defined as $\mathcal{F}^\epsilon = \{f \in \mathcal{F} | \mathcal{L}_p(f) \leq \epsilon\}$. The predictor $f$ can be further decomposed to $h \circ t$, where $t$ is the label-predictive subgraph extractor and $h$ is the subgraph classifier. The set of $t$ can be denoted as $\mathcal{T}$, and what we wish to learn is that $\mathcal{T}(G_i) = S_{G_i}$. To achieve this, it is necessary to foster the diversity of the learned subgraph set. Hence, we impose penalties for overlap among the learned subgraphs.

\begin{figure}
    \centering
    \includegraphics[height=5cm]{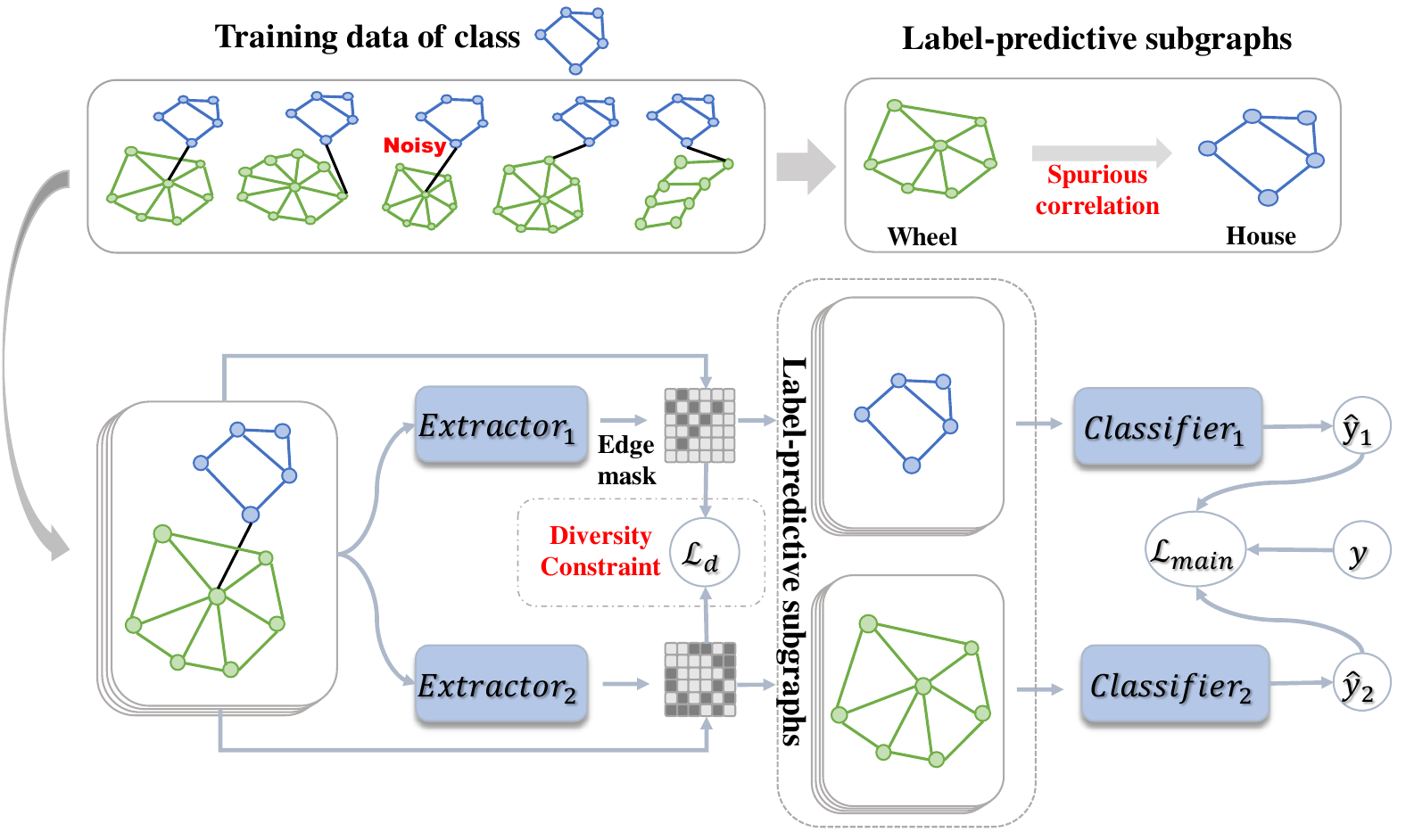}
    \caption{Overall framework of our method when the size of collections is two. The green subgraph (wheel pattern) and the blue subgraph (house pattern) are all label-predictive subgraphs and there exists a strong spurious correlation between these two structrual patterns. We train two models and impose them to attend to different label-predictive subgraph patterns using diversity regularization.}
    \label{fig:framwork}
\end{figure}

\subsection{Model Architecture}
Then, we  describe the architecture of each model within the collection, where each model uniformly shares an identical architectural design and contributes to the learning objective.

\subsubsection{Predictive subgraph extractor} The first part is predictive subgraph extractor. We need to learn a graph mask matrix $M \in \{0,1\}^{N \times N}$ to mask out the predictive subgraph and use it for the subsequent task. It is noteworthy to mention that we are not demand the extractor to extract a invariant subgraph but any predictive subgraph, which can be spurious subgraph or invariant subgraph.

We first encodes the input graph $G$ via a GNN into a set of node representations $\{z_i\}_{v_i\in V}$. For each edge $(v_i, v_j) \in E$, an MLP layer equipped with a sigmoid function is employed to map the concatenated representation of node pair $(z_i, z_j)$ into the masking probability of the edge between them $p_{ij} \in [0, 1]$:
\begin{gather}
Z = \left[\cdots, z_i, \cdots, z_j, \cdots\right]^{\top} = GNN_{mask}(G) \in \mathbb{R}^{n \times d}, \\
p_{ij} = \sigma(MLP_{mask}([z_i, z_j])).
\end{gather}
where $d$ denotes the hidden dimension, $\sigma(\cdot)$ denotes the sigmoid function, and $[\cdot,\cdot]$ denotes the concatenation.
For molecule dataset, the edge representation $e_{ij}$ is also introduced to calculate the $p_{ij}$:
\begin{equation}
    p_{ij} = \sigma(MLP_{mask}([z_i \oplus e_{ij}, z_j \oplus e_{ij}]),
\end{equation}
where $\oplus$ denotes the element-wise sum of vectors.

Consequently, in each forward pass of the training process, we extract a predictive subgraph by sampling from Bernoulli distributions, denoted as  $m_{ij} \sim Bern(p_{ij})$. Due to the inherent non-differentiability of Bernoulli sampling, direct sampling from $Bern(p_{ij})$ can not be optimized. To ensure the gradient of $m_{ij}$ remains calculable, we apply the Gumbel-Sigmoid technique for sampling as:
\begin{equation}
    q_{ij} = Gumbel-Sigmoid(p_{ij}) = \sigma\left(\frac{\log(p_{ij}) + \mathbb{G}}{\tau}\right),
\end{equation}
\begin{equation}
    q'_{ij} = \begin{cases} 
1 & \text{if\quad} q_{ij} > 0.5, \\
0 & \text{if\quad} q_{ij}\leq 0.5,
\end{cases}
\end{equation}
\begin{equation}
        m_{ij} = q'_{ij} + p_{ij} - p_{ij}^\perp,\label{eq: straight}
\end{equation}
where $\mathbb{G} = -\log(-\log(U))$ represents the Gumbel distribution, in which $U\sim Uniform(0,1)$. Given the non-differentiable nature of $q'_{ij}$, we implement the straight-through trick (as delineated in equation (\ref{eq: straight})) to confer a gradient onto $m_{ij}$.  The symbol $\perp$ signifies the cessation of gradient propagation.

The extracted predictive subgraph can be denoted as an induced adjacent matrix $A_P = M \odot A$, where $M$ denotes the learned mask matrix, composed of elements $m_{ij}$. $A$ is the adjacent matrix of original graph $G$, and $\odot$ symbolizes the element-wise multiplication. The subgraph corresponding to $A_P$ is denoted as $G_P$.

\subsubsection{Subgraph encoder and classifier} After obtaining the predictive subgraph $G_P$, we train a GNN model to map the induced subgraph into representation $h_g$, which is fed into the following MLP layer to conduct classification or regression. Formally,
\begin{gather}
    \mathbf{H} = \left[h_1, \cdots, h_n\right]^{\top} = GNN_{feat}(G_P), \\
     h_G = READOUT(\mathbf{H}), \\
     \hat{y} = MLP(h_G) \in \mathcal{Y}.
\end{gather}
\subsubsection{Main task loss} The main task loss can be denoted as:
\begin{equation}
\mathcal{L}_{main} = \mathcal{R}(\hat{Y}, Y),
\end{equation}
where $\mathcal{R}$ represents the task-tailored loss function. For regression tasks, this function implemented with the mean squared error, whereas for classification tasks, it is the cross-entropy function.
\subsection{Diversity via Subgraph Disagreement}

To infuse diversity into the models within the collection, we purposefully advocate for each model to focus on distinct subgraphs. Assume the collection contain $m$ models, the set of predictive masks corresponding to each model can be denoted as
\begin{equation}
    S_{M} = \left\{{M}_1, \cdots ,{M}_m\right\}.
\end{equation}
We apply a jaccard loss as diversity regularizer to penalize the overlapping of each pair of predictive mask in the set:
\begin{equation}
    \mathcal{L}_d = \left.\sum_{i,j}\frac{M_i \cap M_j}{M_i \cup M_j} \middle/ \right.\sum_{i,j}1.
\end{equation}
where $i,j\in\{1,2,\ldots,m\}, i\neq j$ are indices of different models.
\subsection{Learning Objective}
Combining main task loss on each model and diversity regularizer, the total loss of the collection containing $m$ models is defined as:
\begin{equation}
       \mathcal{L}= \frac{1}{m} \sum_{i=1}^m \mathcal{L}_{main}^i + \lambda\mathcal{L}_d,
       \label{eq:loss}
\end{equation}
where $\lambda$ is the hyper-parameter to control the weight of diversity regularization and we set it as 0.5 for all the experiments. 

\subsection{Model Selection}
We employ an OOD validation set for model selection, opting for the model that exhibits the highest validation accuracy across our collection for inference on the test set. The use of an OOD validation set has become a standard practice in contemporary graph-based OOD methods for model selection~\cite{Xiang2023LearningIM, Chen2023DoesIG}. In our experiment, the baseline results were chosen using the same validation set as our method, ensuring  absolute fairness. We have also included the results using the in-distribution (ID) validation set to further demonstrate the efficacy of our methodologies. These results are available at appendix~\ref{appendix: idval}.

\begin{algorithm}[ht]
\caption{Training algorithm for DIVE}\label{alg:cap}

\KwInput{$D_{train}$: training set. $D_{val}$: validation set. $\Theta \longleftarrow \{h_1 \circ t_1, \cdots, h_m \circ t_m\}$: the parameter space of predictor collection.}
\KwOutput{$h^* \circ t^*$: the best preditor}

Initialize the parameter $\Theta$;

\While{not converged}
{
\For{Each Batch $G_{train}^B$ in $D_{train}$}
{
\For{each model  $h_i \circ t_i \in \Theta$}
{
$M_i \longleftarrow  t_i(G_{train}^B)$ \tcp{get subgraph mask for each model}

$G_p^i \longleftarrow M_i \odot G_{train}^B$ \tcp{get the label-predictive subgraph}

$\hat{Y} \longleftarrow h_i(G_p^i)$ \tcp{get the prediction}

$\mathcal{L}_{main}^i \longleftarrow \mathcal{R}(\hat{Y}, Y)$ \tcp{calculate the loss of main task}

$S_M$.append($M_i$) \tcp{add mask to the mask set}
}

$\mathcal{L}_d = \left.\sum_{i,j}\frac{M_i \cap M_j}{M_i \cup M_j} \middle/ \right.\sum_{i,j}1.$ \tcp{calculate the diversity loss}

$\mathcal{L}= \frac{1}{m} \sum_{i=1}^m \mathcal{L}_{main}^i + \lambda\mathcal{L}_d$ \tcp{calculate the total loss}

$\Theta \longleftarrow \Theta - \alpha \Delta_\Theta(\mathcal{L})$ \tcp{update the paramter}
}
}
Select the best predictor $h^* \circ t^*$ using the validation accuracy. 
\end{algorithm}
\section{Experiment}
In this section, we conduct extensive experiments to answer the research questions.  \textbf{(RQ1):} Can our method DIVE achieve better OOD generalization performance against SOTA baselines?  
 \textbf{(RQ2):}Does our method extract subgraphs more accurately than the current methods? 
 \textbf{(RQ3):} Does our regularization really bring diversity to the models in the collection? Is there a model in the collection capable of recognizing invariant structural patterns? 
 \textbf{(RQ4):}What influence does the number of models have on the performance? 
 \textbf{(RQ5):}Does our method robust to the weight of diversity loss?

\subsection{Experimental Setup}
\subsubsection{Datasets}
We employ GOOD and DrugOOD benchmark of graph OOD performance evaluation. 
\begin{itemize}[leftmargin=*]
    \item \textbf{GOOD}~\cite{DBLP:conf/nips/GuiLWJ22}, GOOD is a systematic graph OOD benchmark. It contains two types of distribution shift, covariate shift and concept shift.  In covariate shift, the distribution of input differs. Formally, $P_{train}(G) \neq P_{test}(G)$ and $P_{train}(Y |G) = P_{test}(Y |G)$. While concept shift occurs when the conditional distribution changes as $P_{train}(Y |G)  \neq P_{test(}Y |G)$ and $P_{train}(G) = P_{test}(G)$. Since these two types of distribution shift both contain the spurious correlation, we consider both case in our experiments. We choose four graph-level datasets GOODMotif, GOODHIV, GOODZINC and GOODSST2 to evaluate the graph generalization ability. There are 4 domain in these 4 datasets: basis, scaffold, size and length.  The details of the four datasets can be found at appendix~\ref{appendix: dataset}.
    
    \item \textbf{DrugOOD} is an OOD benchmark for AI-aided drug discovery, offering three environment-splitting strategies: assay, scaffold, and size. These strategies are applied to two measurements, IC50 and EC50. Since the scaffold and size domains overlap with the GOOD benchmark, we focus exclusively on testing the model's generalization ability in the assay domain of IC50 measurement in the DrugOOD benchmark.

\end{itemize}
\subsubsection{Baselines}
We adopted 14 OOD algorithms as baselines including nine graph-specific methods. First, we introduce the general OOD algorithms used for Euclidean data. We utilize two invariant learning baselines based on the invariant prediction assumption. \textbf{IRM}~\cite{arjovsky2019invariant} seeks data representations that perform well across all environments by penalizing features distributions that require different optimal linear classifier for each environment. \textbf{VREx}~\cite{krueger2021out} reducing the variance of risk in test environments  by minimizing the risk variances in training environments. Additionally, we implement two domain adaptation algorithms aimed at minimizing feature discrepancies. \textbf{DANN}~\cite{DBLP:journals/corr/GaninUAGLLML15} adversarially trains a regular classifier and a domain classifier to render features in distinguishable. \textbf{Deep Coral}~\cite{DBLP:conf/eccv/SunS16} minimizes the deviation of covariant matrices from different domains to encourage features similarity across domains. \textbf{GroupDRO}~\cite{DBLP:conf/iclr/SagawaKHL20} addresses the issue of distribution minorities lacking sufficient training through fair optimization, also known as risk interpolation, by explicitly minimizing the loss in the worst training environment. These five methods all need environment labels information.  

To evaluate the performance of current OOD methods specifically for graphs, we include nine graph OOD methods. \textbf{Mixup}~\cite{wang2021mixup} is a data augmentation method designed for graph data. \textbf{DIR}~\cite{wu2022discovering} selects a subset of graph representations as causal rationales and uses interventional data augmentation to create multiple environments.   \textbf{GSAT}~\cite{miao2022interpretable} proposes to build an interpretable graph learning method through attention mechanism and information bottleneck and inject stochasticity into the attention to select label-relevant subgraphs. \textbf{CIGA}~\cite{chen2022learning} proposes an information-theoretic objective to extract the desired invariant subgraphs from the lens of causality. \textbf{GREA}~\cite{liu2022graph} identifies subgraph structures called rationales by environment replacement to create virtual data points to improve generalizability and interpretability. \textbf{CAL}~\cite{sui2022causal} proposes a causal attention learning strategy for graph classification to encourage GNNs to exploit causal features while ignoring the shortcut paths. \textbf{DisC}~\cite{Fan2022DebiasingGN} analyzes the generalization problem of GNNs in a causal view and proposes a disentangling framework for graphs to learn causal and bias substructure. \textbf{MoleOOD}~\cite{yang2022learning} investigates the OOD problem on molecules and designs an environment inference model and a substructure attention model to learn environment-invariant molecular substructures. \textbf{iMoLD}~\cite{Xiang2023LearningIM} introduce a residual vector quantization module that mitigates the over-fitting to training data distributions while preserving the expressivity of encoders.
\subsubsection{Evaluation metrics}
Consistent with the settings in  GOOD benchmark~\cite{DBLP:conf/nips/GuiLWJ22}, We present the accuracy (ACC) for GOODMotif and GOODSST2, and report ROC-AUC score for GOOD-HIV and DrugOOD, as they are binary classification tasks. Additionally, we report the Mean Average Error (MAE) for  GOODZINC dataset since it's regression task.  Our experiments are conducted 10 times using different random seeds. Models are selected based on their performance in the validation dataset, and we report the mean and standard deviations on the test set.

\subsubsection{Implementation Details}
For the training confiuration settings, we employ the Adam optimizer with a weight decay of 0 and a dropout rate of 0.5. The GNN models consist of three convolutional layers. Mean global pooling and the Rectified Linear Unit (ReLU) activation function are utilized, with a hidden layer dimension of 300. The batch size is set to 32, the maximum number of epochs is 300, and the initial learning rate is 1e-3. During the training process, all models are trained until convergence is achieved. In terms of computational resources, we typically employ one NVIDIA GeForce RTX 3090 for each individual experiment. For the hyperparamtere selection, indeed, one main advantage of our method is that our method does not need laborious hyper-paramter tuning. Our method only has one hyper-parameter: the weight $\lambda$ of diveristy loss, and we set is as 0.5 for all settings.

\subsection{Result Comparison and Analysis}

\begin{table*}[ht]
\centering
\caption{Results of synthetic datasets and text-attributed datasets. We use the accuracy ACC (\%) as the evaluation metric. The best and the second-best results are highlighted in \textbf{bold} and \underline{underline} respectively. DIVE-N means that the size of the predictor collection is N.}
\begin{tabular}{l|cccccc}
\toprule
Method & \multicolumn{4}{c}{GOOD-Motif $\uparrow$} & \multicolumn{2}{c}{GOOD-SST2 $\uparrow$} \\
\cmidrule(r){2-5} \cmidrule(r){6-7} 

 & \multicolumn{2}{c}{basis} & \multicolumn{2}{c}{size} & \multicolumn{2}{c}{length}\\
 & covariate & concept & covariate &concept & covariate & concept \\
\midrule
ERM & 63.80\scriptsize(10.36) & 81.31\scriptsize(0.69) & 53.46\scriptsize(4.08) & {70.83\scriptsize(0.79)}& 80.52\scriptsize(1.13) & 72.92\scriptsize(1.10) \\
IRM & 59.93\scriptsize(11.46) & 80.37\scriptsize(0.80) & 53.68\scriptsize(4.11) & 70.15\scriptsize(0.64) & 80.75\scriptsize(1.17) & \textbf{77.45\scriptsize(2.37)}\\
VREx & 66.53\scriptsize(4.04) & 81.34\scriptsize(0.75) & 54.47\scriptsize(3.42) & 70.58\scriptsize(1.16) & 80.20\scriptsize(1.39) & 72.92\scriptsize(0.95)\\
GroupDRO & 61.96\scriptsize(8.27) & 81.00\scriptsize(0.60) & 51.69\scriptsize(2.22) & 70.35\scriptsize(0.40) & {81.67\scriptsize(0.45)}& 72.51\scriptsize(0.79)\\
Coral & 66.23\scriptsize(9.01) & 81.47\scriptsize(0.49) & 53.71\scriptsize(2.75) & 70.52\scriptsize(0.59) & 78.94\scriptsize(1.22) & 72.98\scriptsize(0.46) \\ 
DANN & 51.54\scriptsize(7.28) & 81.43\scriptsize(0.60) & 51.86\scriptsize(2.44) & 70.74\scriptsize(0.65) & 80.53\scriptsize(1.40) & 74.10\scriptsize(1.49)\\\midrule
Mixup & {69.67\scriptsize{(5.86)}} & 77.64\scriptsize{(0.58)} & 51.31\scriptsize{(2.56)} & 68.21\scriptsize{(0.89)} & 80.77\scriptsize{(1.03)} & 72.57\scriptsize{(0.76)} \\
DIR & 39.99\scriptsize{(5.50)} & {82.96\scriptsize{(4.47)}} & 44.83\scriptsize{(4.00)} & 54.96\scriptsize{(9.32)} & 81.55\scriptsize{(1.06)} & 67.98\scriptsize{(3.07)} \\
GSAT & 55.13\scriptsize{(5.41)} & 75.30\scriptsize{(1.57)} & {60.76\scriptsize{(5.94)}} & 59.00\scriptsize{(3.42)} & 81.49\scriptsize{(0.76)} & 74.54\scriptsize{(1.40)} \\
CIGA & 67.15\scriptsize(8.19) & 77.48\scriptsize(2.54) & 54.42\scriptsize(3.11) & 70.65\scriptsize(4.81)  & 80.44\scriptsize{(1.24)} & 71.18\scriptsize{(1.91)}  \\ \midrule
DIVE-2 & \underline{84.05\scriptsize(5.25)} &  \textbf{92.01\scriptsize(1.47)} & \underline{73.73\scriptsize(2.41)} & \underline{71.02\scriptsize(1.12)} & {83.08\scriptsize(1.01)} & {75.74\scriptsize(1.14)} \\ 
DIVE-3 & \textbf{85.77\scriptsize(4.32)}& \underline{89.05\scriptsize(2.34)} &  \textbf{75.05\scriptsize(3.36)}& \textbf{72.01\scriptsize(4.15)} & \underline{83.44\scriptsize(1.21)} & \underline{75.89\scriptsize(1.33)} \\
DIVE-4 & 80.23\scriptsize(3.45) & 88.90\scriptsize(2.88) & 68.77\scriptsize(4.64) & 70.10\scriptsize(4.71) &  \textbf{83.71\scriptsize(1.33)} & 75.67\scriptsize(1.28) \\
$\Delta$\emph{Improve.} & 23.10 \% & 10.90\% & 23.46\% & 1.66\% & 2.49\% & -2.01\%   \\ \midrule
w/o diversity  & 70.13\scriptsize(5.41)   & 78.30\scriptsize(1.01) & 61.71\scriptsize(2.09)  & 58.00\scriptsize(1.04) &  81.01\scriptsize(0.75)  & 73.24\scriptsize(1.21)    \\
\bottomrule
\end{tabular}
\label{tb: motif}
\end{table*}

\begin{table*}[htbp]
\centering
\caption{Overall performance of molecular datasets. We compare the performance of 14 methods on three molecular datasets. The results of  GOODHIV and DrugOOD are reported in terms of ROC-AUC. The results of GOODZINC are reported using MAE. - denotes abnormal results caused by under-fitting declared in  GOOD benchmark, and / denotes that the method cannot be applied to this dataset.}
\label{tab:example}
\begin{tabular}{lccccccccc}
\toprule
Method & \multicolumn{4}{c}{GOOD-ZINC $\downarrow$} & \multicolumn{4}{c}{GOOD-HIV$\uparrow$} & DrugOOD$\uparrow$ \\
\cmidrule(r){2-5} \cmidrule(r){6-9} \cmidrule{10-10}
 & \multicolumn{2}{c}{scaffold} & \multicolumn{2}{c}{size} & \multicolumn{2}{c}{scaffold} & \multicolumn{2}{c}{size} & assay \\
 & covariate & concept & covariate & concept & covariate & concept & covariate & concept & covariate \\
\midrule
ERM & 0.1802\scriptsize(0.0174) & 0.1301\scriptsize(0.0052) & 0.2319\scriptsize(0.0072) & 0.1325\scriptsize(0.0085) & 69.55\scriptsize(2.39) & 72.48\scriptsize(1.26) & 59.19\scriptsize(2.29) & 61.91\scriptsize(2.29) & 71.63\scriptsize(0.76)\\
IRM & 0.2164\scriptsize(0.0160) & 0.1339\scriptsize(0.0043) & 0.6984\scriptsize(0.2809) & 0.1336\scriptsize(0.0055) & 70.17\scriptsize(2.78) & 71.78\scriptsize(1.37) & 59.94\scriptsize(1.59) & -\scriptsize(-) & 71.15\scriptsize(0.57) \\
VREx & 0.1815\scriptsize(0.0154) & 0.1287\scriptsize(0.0053) & 0.2270\scriptsize(0.0136) & 0.1311\scriptsize(0.0067) & 69.34\scriptsize(3.54) & 72.21\scriptsize(1.42) & 58.49\scriptsize(2.28) & 61.21\scriptsize(2.00) & 72.32\scriptsize(0.58) \\
GroupDRO & 0.1870\scriptsize(0.0128) & 0.1323\scriptsize(0.0041) & 0.2377\scriptsize(0.0147) & 0.1333\scriptsize(0.0064) & 68.15\scriptsize(2.84) & 71.48\scriptsize(1.27) & 57.75\scriptsize(2.86) & 59.77\scriptsize(1.95) & 71.57\scriptsize(0.48)\\
Coral & 0.1769\scriptsize(0.0152) & 0.1303\scriptsize(0.0057) & 0.2292\scriptsize(0.0090) & 0.1261\scriptsize(0.0060) & 70.69\scriptsize(2.25) & 72.96\scriptsize(1.06) & 59.39\scriptsize(2.90) & 60.29\scriptsize(2.50) &  71.28\scriptsize(0.91) \\ 
DANN & 0.1746\scriptsize(0.0084) & 0.1269\scriptsize(0.0042) & 0.2326\scriptsize(0.0140) & 0.1348\scriptsize(0.0091) & 69.43\scriptsize(2.42) & 71.70\scriptsize(0.90) & 62.38\scriptsize(2.65) & 65.15\scriptsize(3.13) & 69.84\scriptsize(1.41) \\ \midrule
Mixup & 0.2066\scriptsize(0.0123) & 0.1391\scriptsize(0.0071) & 0.2531\scriptsize(0.0150) & 0.1547\scriptsize(0.0082) & 70.65\scriptsize(1.86) & 71.89\scriptsize(1.73) & 59.11\scriptsize(3.11) & 62.80\scriptsize(2.43) & 71.49\scriptsize(1.08)\\
DIR & 0.3682\scriptsize(0.0639) & 0.2543\scriptsize(0.0454) & 0.4578\scriptsize(0.0412) & 0.3146\scriptsize(0.1225) & 68.44\scriptsize(2.51) & 71.40\scriptsize(1.48) & 57.67\scriptsize(3.75) & 74.39\scriptsize(1.45) & 69.84\scriptsize(1.41)  \\
GSAT & 0.1418\scriptsize(0.0077) & 0.1066\scriptsize(0.0046) & 0.2101\scriptsize(0.0095) & 0.1038\scriptsize(0.0030) & 70.07\scriptsize(1.76) & 72.51\scriptsize(0.97) & 60.73\scriptsize(2.39) & 56.96\scriptsize(1.76) & 70.59\scriptsize(0.43) \\
CIGA & \multicolumn{4}{c}{-} & 69.40\scriptsize(2.39) & 70.79\scriptsize(1.55) & 61.81\scriptsize(1.68) & 72.80\scriptsize(1.35) & 71.86\scriptsize(1.37)\\
GREA & 0.1691\scriptsize(0.0159) & 0.1157\scriptsize(0.0084) & 0.2100\scriptsize(0.0081) & 0.1273\scriptsize(0.0044) & 71.98\scriptsize(2.87) & 70.76\scriptsize(1.16) & 60.11\scriptsize(1.07) & 60.96\scriptsize(1.55) & 70.23\scriptsize(1.17) \\
CAL & \multicolumn{4}{c}{-} & 69.12 \scriptsize(1.10) & 72.49\scriptsize(1.05) & 59.34\scriptsize(2.14) & 56.16\scriptsize(4.73) & 70.09\scriptsize(1.03) \\
DisC & \multicolumn{4}{c}{-} & 58.85\scriptsize(7.26) &64.82\scriptsize(6.78) &49.33\scriptsize(3.84) &74.11\scriptsize(6.69)  & 61.40\scriptsize(2.56)\\
MoleOOD & 0.2752\scriptsize(0.0288) & 0.1996\scriptsize(0.0136) & 0.3468\scriptsize(0.0366) & 0.2275\scriptsize(0.2183) & 69.39\scriptsize(3.43)& 69.08\scriptsize(1.35) &58.63\scriptsize(1.78) &55.90\scriptsize(4.93) & 71.62\scriptsize(0.52) \\
iMoLD & 0.1410\scriptsize(0.0054) & 0.1014\scriptsize(0.0040) & 0.1863\scriptsize(0.0139) & 0.1029\scriptsize(0.0067) & \underline{72.93\scriptsize(2.29)} & \textbf{74.32\scriptsize(1.63)} &62.86\scriptsize(2.58) &77.43\scriptsize(1.32) &71.86\scriptsize(1.37)\\  \midrule
DIVE-2 & 0.1279\scriptsize(0.0045) & 0.0674\scriptsize(0.0051) & 0.1305\scriptsize(0.0038) & 0.0531\scriptsize(0.0061) & 72.87\scriptsize(1.33) & 72.51\scriptsize(2.56) & \underline{63.80\scriptsize(2.21)}& \textbf{78.09\scriptsize(3.12)} &  \underline{73.20\scriptsize(1.33)} \\

DIVE-3  & \underline{0.1245\scriptsize(0.0043)} & \underline{0.0644\scriptsize(0.0061)} & \textbf{0.1250\scriptsize(0.0071)}& \underline{0.0529\scriptsize(0.0054)} & \textbf{73.33\scriptsize(1.65)} & \underline{73.30\scriptsize(1.45)} & \textbf{64.03\scriptsize(2.47)} & \underline{77.78\scriptsize(4.54)} & \textbf{73.55\scriptsize(1.32)}  \\
DIVE-4  & \textbf{0.1201\scriptsize(0.0038)} & \textbf{0.0581\scriptsize(0.0042)} & \underline{0.1256\scriptsize(0.0040)} & \textbf{0.0501\scriptsize(0.0041)} &  69.89\scriptsize(1.93) & 71.56\scriptsize(1.98) & 62.88\scriptsize(1.90) & 74.22\scriptsize(3.43) &   72.67\scriptsize(1.11)  \\
$\Delta$\emph{Improve.} & 14.80\% & 42.70\% & 32.90\%& 51.31\% & 0.54\% & -1.37\% & 1.86\%  & 0.84\% & 2.29\% \\ \midrule
w/o diversity  & 0.1518\scriptsize(0.0072)  & 0.1166\scriptsize(0.0045)  & 0.2204\scriptsize(0.0044)  & 0.1236\scriptsize(0.0030)  & 70.07\scriptsize(1.76)  &  71.36\scriptsize(0.97)  & 60.73\scriptsize(2.39)  & 56.96\scriptsize(1.76) & 70.33\scriptsize(2.15)   \\
\bottomrule
\end{tabular}
\label{tb: molecular}
\end{table*}

\begin{figure*}[!ht]
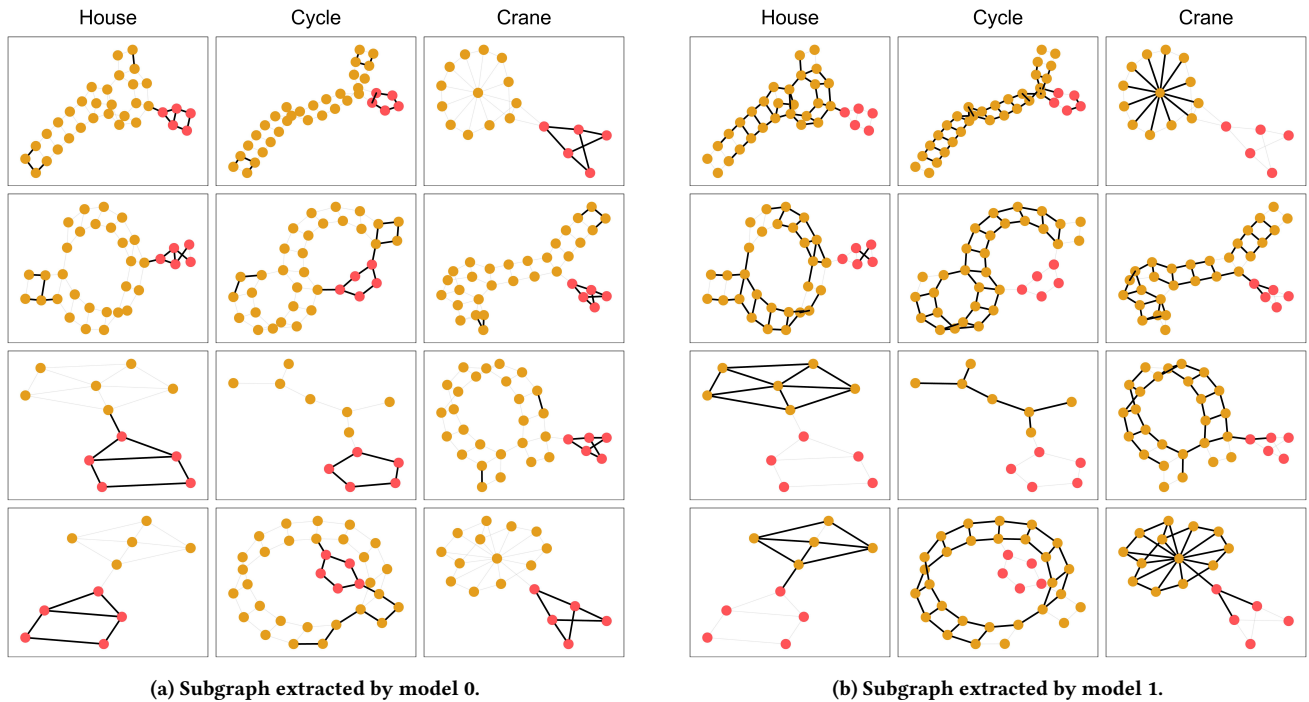

    \centering
    \begin{subfigure}[b]{0.49\textwidth}
        \centering
        \includegraphics[width=0.95\textwidth]{fig/experiment/model1_att.pdf} 
        \caption{Subgraph extracted by model 0.}
        \label{fig:model1}
    \end{subfigure}\hfill
    \begin{subfigure}[b]{0.49\textwidth}
        \centering
        \includegraphics[width=0.95\textwidth]{fig/experiment/model2_att.pdf} 
        \caption{Subgraph extracted by model 1.}
        \label{fig:model2}
    \end{subfigure}
    \caption{Visualization of the subgraph masks generated by different models in the collections. We train two model using our algorithm on  GOODMotif dataset (basis-concept setting) and visualize the subgraph extracted by each model on the test set. Nodes colored pink are ground-truth subgraph nodes and each column represents a graph class. Subfigures (a) and (b), located as the identical position, correspond to each other and represent the same graph instance. It can be observed that model 0 attends to the correct subgrah while model 1 attends to the spurious one.}
    \label{fig:edge-masks}
\end{figure*}

\subsubsection{Main task results (RQ1)}
In this section, our goal is to address Research Question 1 (RQ1) by conducting a comparative analysis of our approach, DIVE, against various baseline methodologies. The performance of DIVE in contrast to the current state-of-the-art (SOTA) methods is delineated in Tables \ref{tb: molecular} and \ref{tb: motif}. DIVE achieves superior outcomes in 13 out of 15 scenarios across five datasets. In the two remaining scenarios, it secures the second-best positions. It is noteworthy that the Invariant Risk Minimization (IRM) method requires environmental labels for each instance during training. Consequently, excluding methods necessitating environmental labels, our approach secures the top position in 14 out of 15 cases. A significant enhancement is observed in the datasets GOOD-Motif and GOOD-ZINC. Specifically, in GOOD-ZINC dataset, DIVE marks an impressive 51.31\% improvement in the size-concept scenario, which may imply that our method is more competitive on regression datsets.  Unlike most existing methods that excel in limited scenarios but experience substantial performance declines in others, DIVE consistently demonstrates top-tier performance across a majority of senarios. This underscores the efficacy of DIVE in extracting invariant subgraphs.

Conversely, subgraph-mixup methodologies, including mixup, DIR, GREA, and Disc, generally underperform across the board, frequently yielding results inferior to Empirical Risk Minimization (ERM). This suggests that current subgraph-mixup approaches fail to accurately isolate invariant subgraphs, and the incorporation of mixup can intensify the issue of spurious correlations if the extracted subgraph harbor spurious information. Furthermore, the representative information bottleneck method, GSAT, fails to achieve satisfactory outcomes across these datasets. This indicates a limitation of the information bottleneck technique in distinguishing between invariant and merely predictive features, thereby rendering it ineffective in addressing spurious correlations.

Additionally, MoleOOD, which necessitates the inference of environmental labels, also shows poor performance on three datasets. This highlights the complexities and challenges associated with inferring environment labels, where inaccuracies in inferred labels can detrimentally affect the overall results.

\begin{figure}
    \centering
    \includegraphics[width=\linewidth]{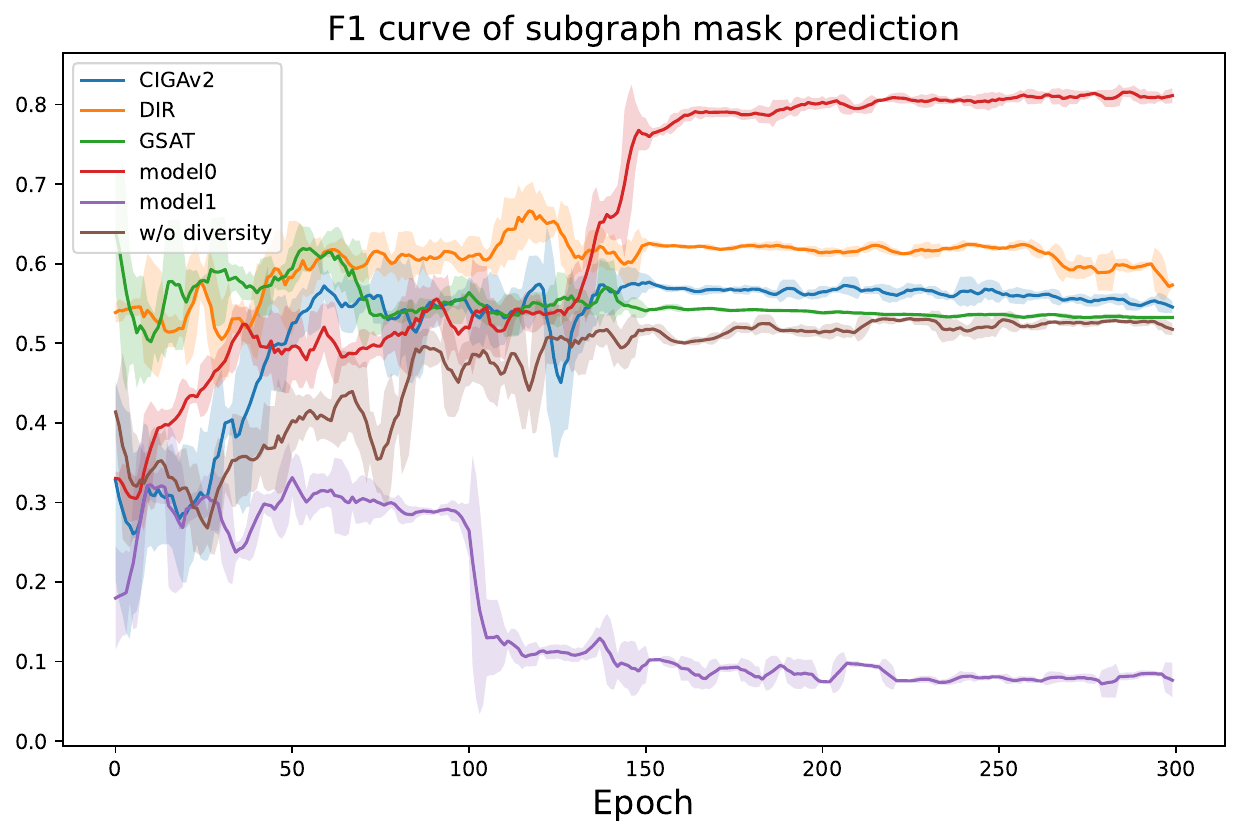}
    \caption{F1 curve of the subgraph mask prediction. For each method, we run the experiment for 5 times and the shadowed area represents standard deviation.}
    \label{fig: f1}
\end{figure}

\subsubsection{The inadequacy of the current method in precisely extracting subgraphs (RQ2).}
To directly compare our method with current method that based on subgraph extraction in terms of invariant subgraph extraction ability, we conducted comparative analysis with the DIR, GSAT, and CIGA methods, evaluating the efficacy of true subgraph extraction. Analogous to our technique, these three methodologies learn an adjacency matrix mask to extract the subgraph. Since  GOOD-Motif dataset annotates  the ground-truth subgraph mask for each graph instance, we conduct the experiment on this dataset. We computed the F1 score of subgraph mask prediction for each graph instance within the dataset and calculate the mean F1 score.  We train a collection with  two models incorporating  the diversity regularization. As illustrated in Figure \ref{fig: f1}, the majority of current methods achieve at most a F1 score of 0.6.  As elucidated in the introduction, this signifies the current methods' inability to accurately extract the correct subgraphs. Under these circumstances, employing the mixup technique strengthen the spurious correlation. whereas our approach surpasses this threshold, attaining a F1 score exceeding 0.8 for subgraph mask prediction. In the absence of diversity regularization, our method's performance diminishes, achieving a F1 score around 0.5. However, with diversity regularization, there is a notable improvement in the F1 score of one model within the ensemble, consistently rising to surpass 0.8. Conversely, due to the influence of diversity regularization, the other model becomes predisposed towards the spurious aspects, resulting in a progressive decline in its F1 score to 0.1.
\subsubsection{Results on diversity of the collections (RQ3)} 
\begin{figure}[h!]
    \centering
    \begin{subfigure}[b]{0.49\linewidth}
        \centering
        \includegraphics[width=0.95\linewidth]{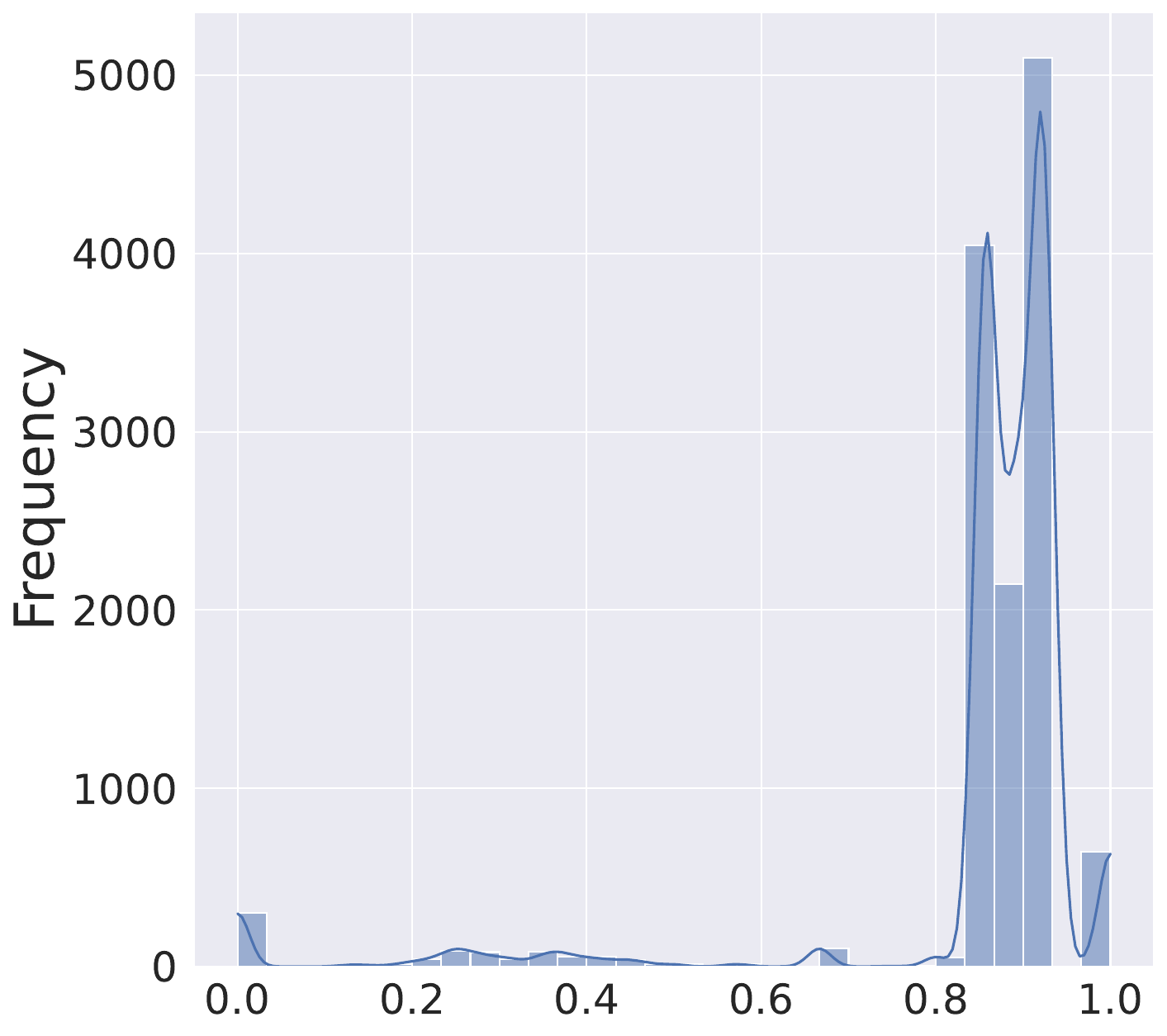} 
        \caption{precision of model 0}
        \label{fig:sub1}
    \end{subfigure}\hfill
    \begin{subfigure}[b]{0.49\linewidth}
        \centering
        \includegraphics[width=0.95\linewidth]{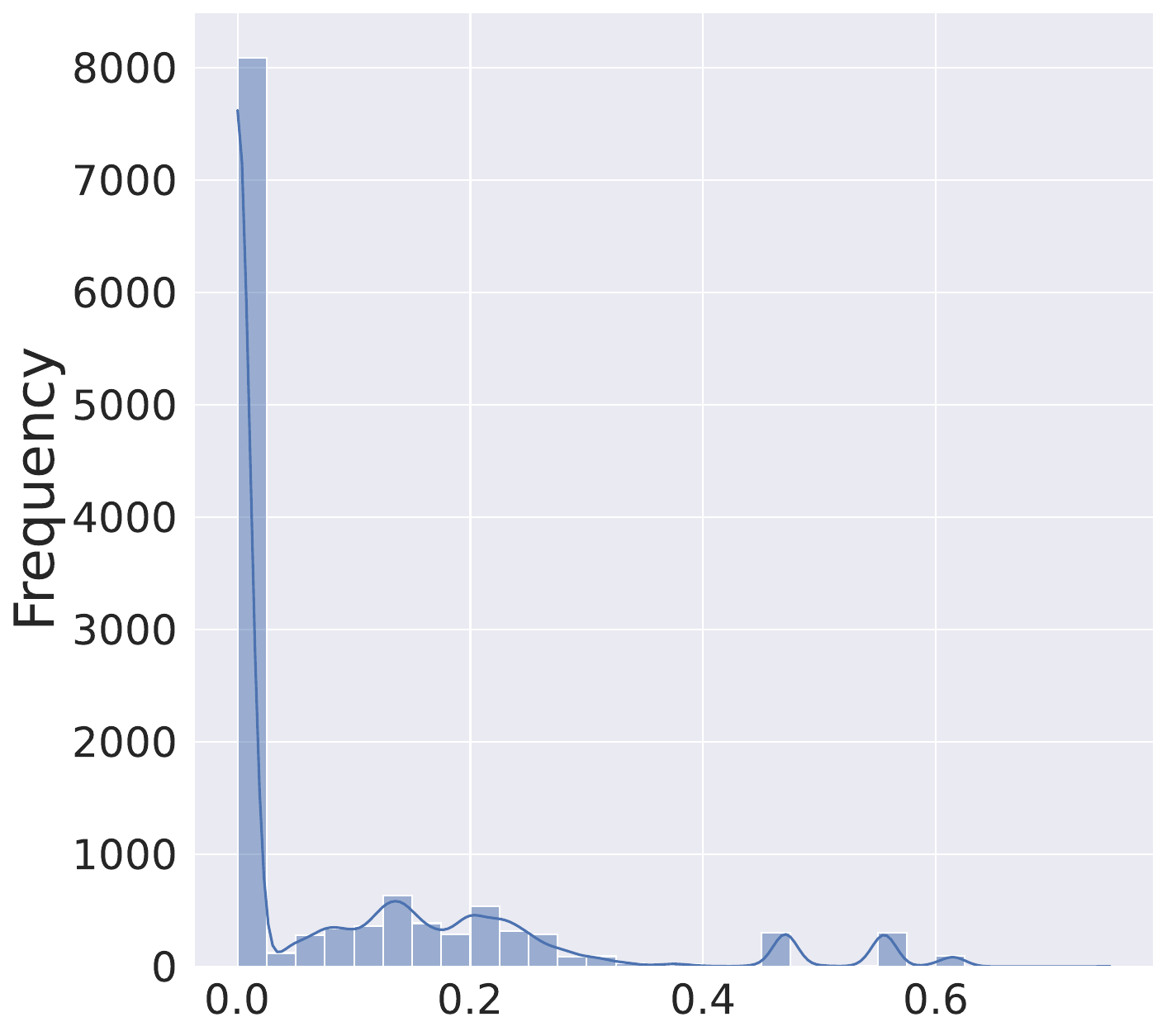} 
        \caption{precision of model 1}
        \label{fig:sub2}
    \end{subfigure}

    \begin{subfigure}[b]{0.49\linewidth}
        \includegraphics[width=\linewidth]{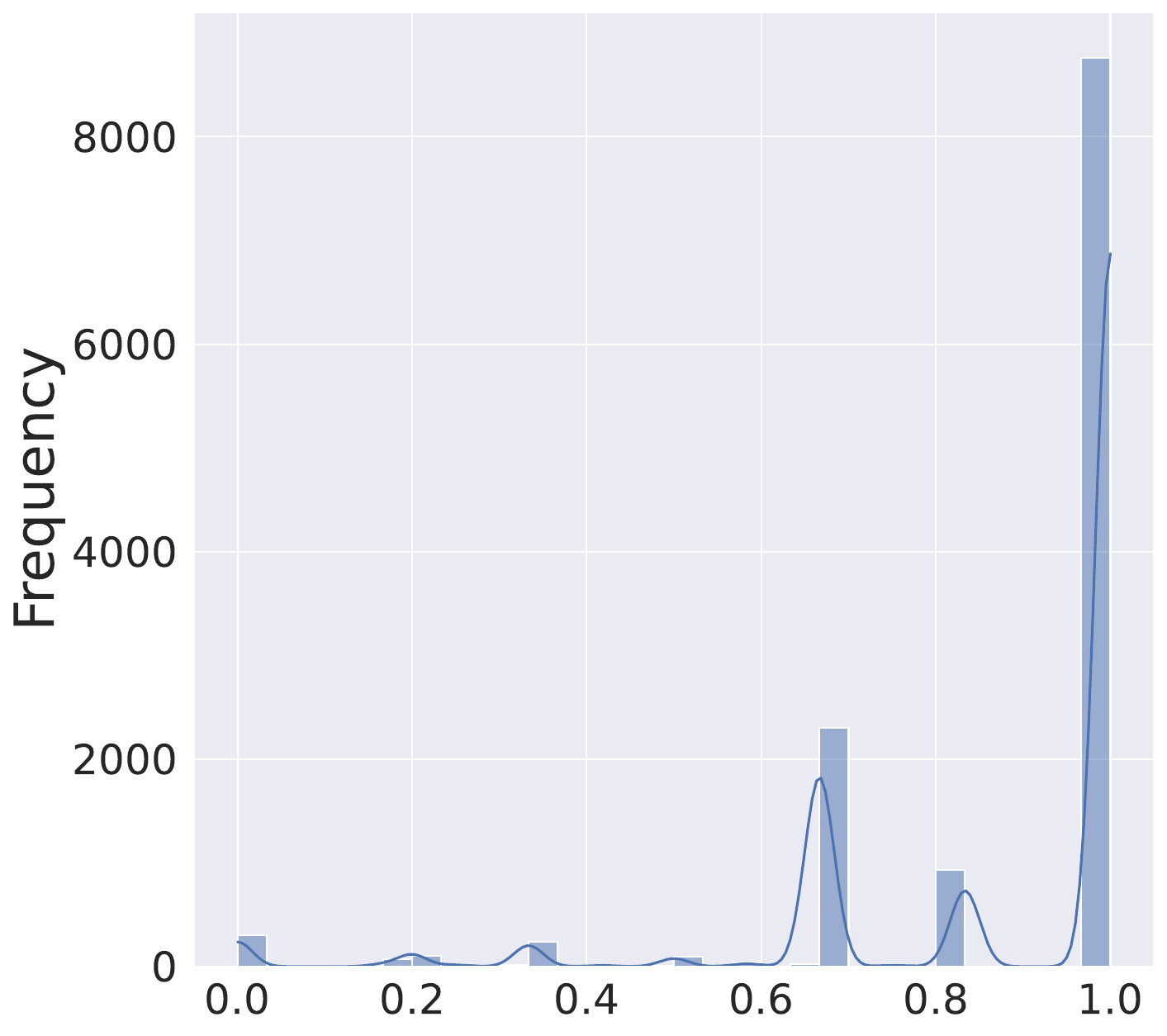}
        \caption{recall of model 0}
        \label{fig:sub3}
    \end{subfigure}
    \hfill
    \begin{subfigure}[b]{0.49\linewidth}
        \includegraphics[width=\linewidth]{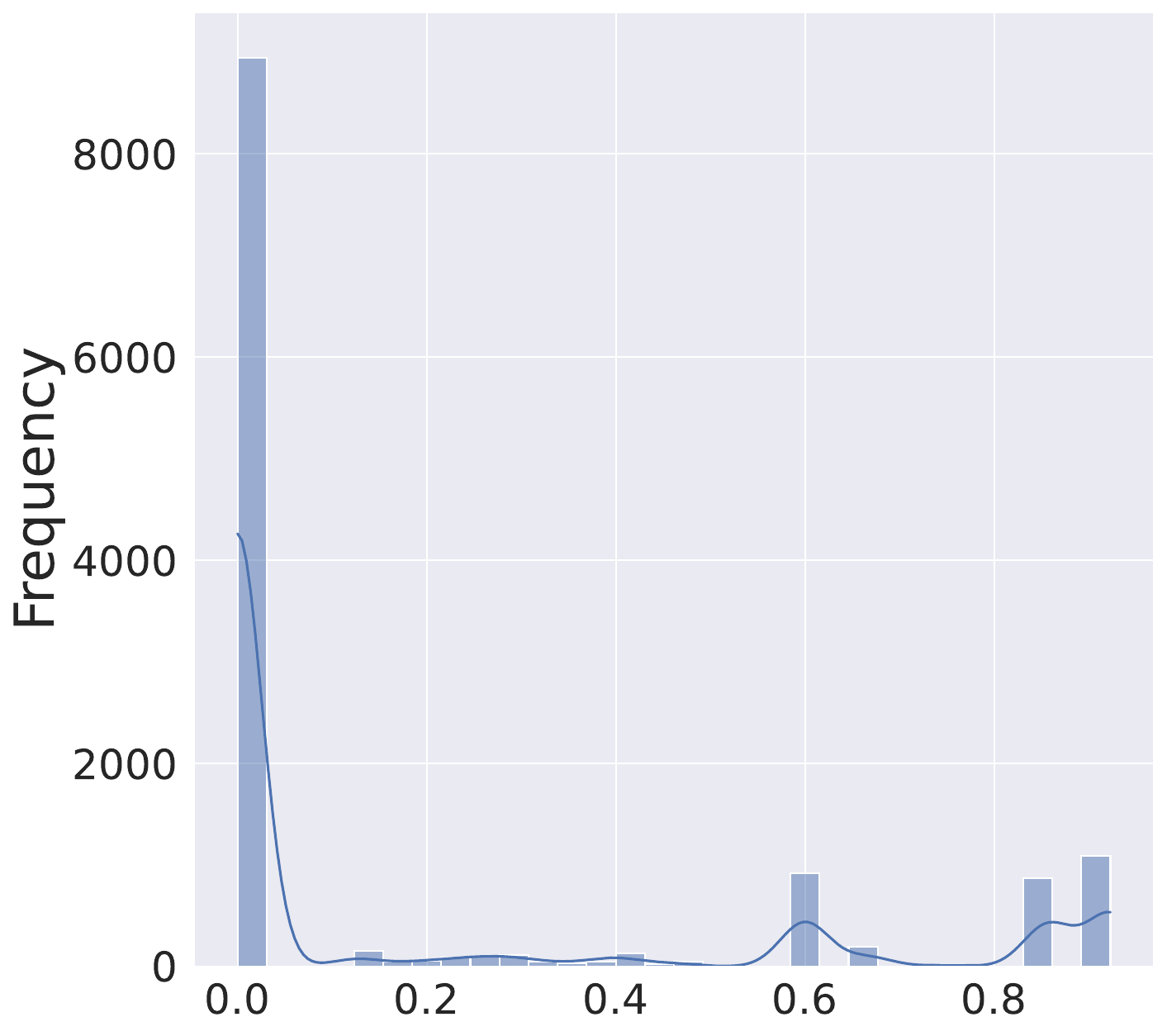}
        \caption{recall of model 1}
        \label{fig:sub4}
    \end{subfigure}       
    
    \caption{Distribution of the subgraph mask precision and recall of different models in the collection. }
    \label{fig:distribution}
\end{figure}

We present a visualization of the predictive subgraph identified by DIVE within GOODMotif (basis-concept scenario) test set. In this scenario, each graph in the dataset is synthesized by integrating a base graph (ladder, wheel, tree) with a motif (tree, cycle, crane), where the motif exclusively determines the label of the graph. In the training dataset, the base graph exhibits a high degree of spurious correlation with the label. We demonstrate the subgraph learned on  GOODMotif dataset's test set to ascertain if the models in the collection can focus on different subgraphs. We visualize four subgraph masks produced by models in the collection for each class. As illustrated in Figure~\ref{fig:edge-masks}, the two models in the collection concentrate on distinct parts. Model 0 focuses on the motif part, which is crucial for determining the label. Conversely, model 1 primarily focuses on the spurious part and is likely to make incorrect predictions on the test set, as the spurious correlation is present only in the training set and not in the test set.Additionally, we display the distribution of subgraph mask precision and recall for various models within the collection as Figure~\ref{fig:distribution}. It becomes evident that model 0 exhibits markedly higher precision and recall compared to model 1. The bulk of precision and recall values for model 1 are concentrated near zero, indicating that model 1 is overly influenced by the base graph and disregards the critical subgraph. In contrast, model 0 displays precisely the opposite behavior because of the subgraph diversity regularization. We also present the metric curves for different models in the collection, which is detailed at appendix \ref{appendix: curve}.

\subsubsection{Ablation study (RQ3)}
We carry out the ablation study to examine the discrepancy in performance between our approach with and without the implementation of diversity regularization. As evidenced by the final row of Tables \ref{tb: motif} and \ref{tb: molecular}, the absence of diversity regularization leads to a considerable decline in performance across all scenarios of every dataset. This indicates that diversity regularization is essential for our methods to attain superior generalization capabilities. 

\subsubsection{The impact of size of the collection (RQ4)} We report the results when the size of collections is [2,3,4] in table~\ref{tb: motif} and table~\ref{tb: molecular}. It was observed that on GOODMotif dataset, either 2 or 3 models suffice for our methodology to attain commendable out-of-distribution (OOD) performance, whereas employing 4 models leads to a decline in performance. This can be attributed to the fact that GOODMotif dataset is synthetic, with each graph instance only being a composite of a base graph and a motif, hence limiting the structural pattern variability. Most optimal results are achieved with 3 models, as the basic graph structure (spurious part) can be bifurcated into two segments, exemplified by the division of the wheel graph into the edges of the wheel's outer rim and the hub's edges. However, an increase in the number of models does not contribute to the identification of more predictive structural patterns. On the real datasets, it was observed that our methodology attains optimal performance with a collection size of 4 for  GOODZINC dataset. Conversely, for other datasets, a collection of 3 models suffices to reach optimal outcomes. This discrepancy can likely be attributed to the considerable size of GOODZINC, which encompasses approximately 250,000 samples, in contrast to other datasets that contain no more than 100,000 samples each. Consequently, GOODZINC dataset may possess more predictive structural patterns, rendering additional models beneficial in uncovering more of these predictive patterns.

\subsubsection{Hyperparameter analysis (RQ5)}
\begin{figure}[h!]
    \centering
    \begin{subfigure}[b]{0.49\linewidth}
        \centering
        \includegraphics[width=0.95\linewidth]{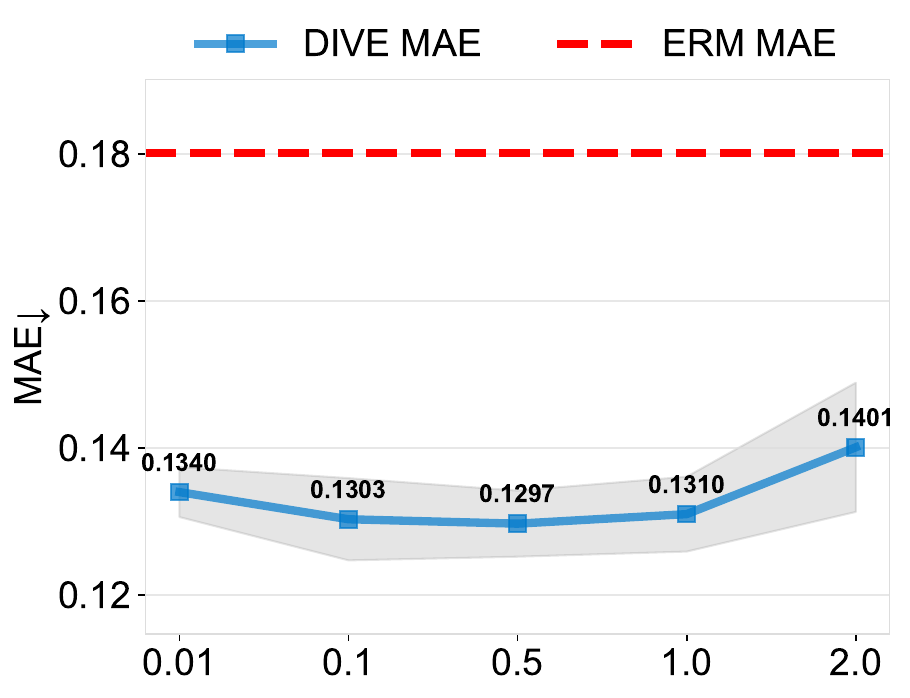} 
        \caption{Performance on GOODZINC}
        \label{fig:model1_}
    \end{subfigure}\hfill
    \begin{subfigure}[b]{0.49\linewidth}
        \centering
        \includegraphics[width=0.95\linewidth]{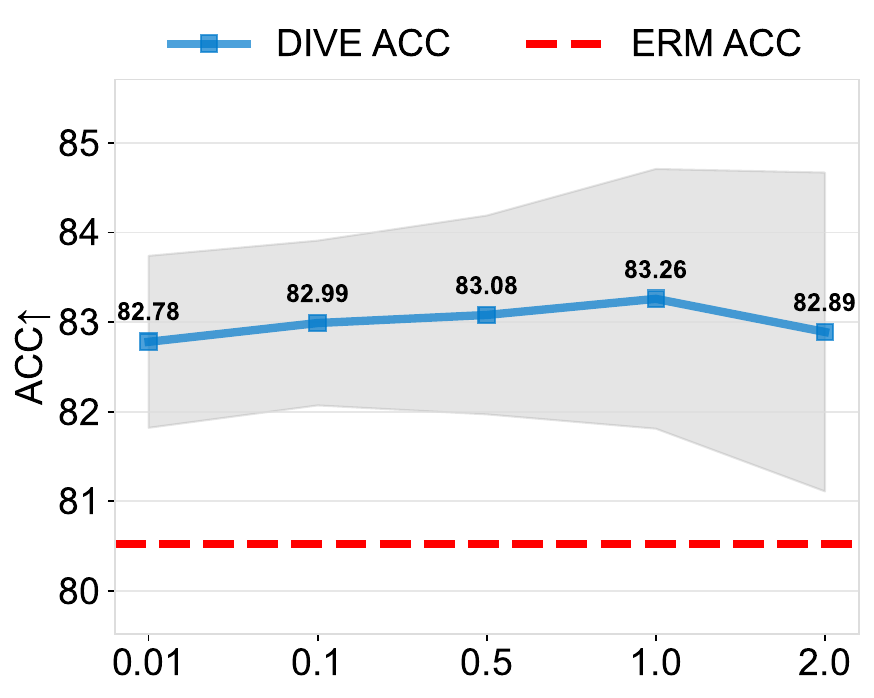} 
        \caption{Performance on GOODSST2}
        \label{fig:model2_}
    \end{subfigure}
    \caption{Performance using  different $\lambda$ on GOODZINC and GOODSST2. We conduct the experiment 5 times for each $\lambda$ and the grey shaded area represents standard deviation.}
    \label{fig:lambda}
\end{figure}

Without losing the generality, we conduct the sensitivity analysis on two datasets: GOODZINC and GOODSST2. We show our model's performance when the $\lambda$ is [0.01, 0.1, 0.5, 1, 2]. Figure \ref{fig:lambda} shows that the performance of our methods is insensitive to the hyperparameter $\lambda$ in Eq. \ref{eq:loss}.

\section{Conclusion}
In this study, we introduce a new learning paradigm named DIVE, designed to address the graph out-of-distribution challenge. This approach involves the development of an ensemble of diverse models capable of focusing on all label-predictive subgraphs, thereby reducing the impact of simplicity bias during training. We employ a subgraph diversity regularization technique to promote the variation in structural patterns recognized by the models. Comprehensive experiments conducted on one synthetic dataset and four real-world datasets underscore the exceptional performance of DIVE.


\begin{acks}
This work is sponsored by the National Key Research and Development Program (2023YFC3305203), and the National Natural Science Foundation of China (NSFC) (62206291, 62141608). 
\end{acks}

\balance
\bibliographystyle{ACM-Reference-Format}
\bibliography{sample-base}

\appendix

\begin{table*}[!h]
\begin{tabular}{ccccccccccc}
\toprule
\multirow{3}{*}{Methods} & \multicolumn{4}{c}{GOOD-Motif}                       & \multicolumn{4}{c}{GOOD-ZINC}                                 & \multicolumn{2}{c}{GOOD-SST2} \\ \cline{2-11} 
                         & \multicolumn{2}{c}{basis} & \multicolumn{2}{c}{size} & \multicolumn{2}{c}{scaffold}  & \multicolumn{2}{c}{size}    & \multicolumn{2}{c}{length}    \\
                         & covariate    & concept    & covariate   & concept    & covariate     & concept       & covariate     & concept       & covariate     & concept       \\ \midrule
ERM                      & 69.97\scriptsize(1.94)   & 80.87\scriptsize(0.65) & 51.28\scriptsize(1.94)  & 69.41\scriptsize(0.91) & 0.1825\scriptsize(0.0129) & 0.1328\scriptsize(0.0060) & 0.2569\scriptsize(0.0138) & 0.1418\scriptsize(0.0057) & 77.76\scriptsize(1.14)    & 67.26\scriptsize(0.05)    \\
GSAT                     & 63.33\scriptsize(5.34)   & 76.43\scriptsize(2.13) & 43.20\scriptsize(6.45)  & 49.01\scriptsize(2.66) & 0.1634\scriptsize(0.0234) & 0.1342\scriptsize(0.0058) & 0.2418\scriptsize(0.0098) & 0.1309\scriptsize(0.0116) & 72.56\scriptsize(2.77)    & 61.45\scriptsize(2.56)    \\
DIR                      & 59.08\scriptsize(14.23)  & 67.57\scriptsize(2.71) & 42.61\scriptsize(1.31)  & 53.21\scriptsize(4.03) & 0.6155\scriptsize(0.0589) & 0.3883\scriptsize(0.1019) & 0.6011\scriptsize(0.0147) & 0.3130\scriptsize(0.0747) & 74.76\scriptsize(2.31)    & 63.61\scriptsize(1.32)    \\
CIGA                     & 55.12\scriptsize(10.12)  & 61.58\scriptsize(3.12) & 44.69\scriptsize(4.35)  & 51.45\scriptsize(5.69) & -             & -             & -             &               & 63.78\scriptsize(2.02)    & 56.92\scriptsize(2.54)    \\
DIVE-2                   & 81.50\scriptsize(6.23)   & 89.54\scriptsize(1.34) & 63.66\scriptsize(1.54)  & 70.12\scriptsize(2.30) & 0.1290\scriptsize(0.0433) & 0.0697\scriptsize(0.0688) & 0.1328\scriptsize(0.0071) & 0.0631\scriptsize(0.0081) & 82.38\scriptsize(1.45)    & 72.78\scriptsize(1.22)    \\ \bottomrule
\end{tabular}
\caption{The results on ID validation set}

\end{table*}

\section{The details of datasets}
\label{appendix: dataset}
The details of the four datasets from the GOOD benchmark are as follows:
\begin{itemize}
        \item \textbf{GOODMotif} is a synthetic datasets motivated by Spurious-Motif and is designed for structural shifts. Each graph in the dataset is generated by connecting a base graph and a motif, and the label is determined by the motif solely. The graphs are generated using five label irrelevant base graphs and three label determinant motifs(house, cycle, and crane). The base graph type and the size is utilized to split the domain.
        \item \textbf{GOODHIV} is a small-scale, real-world molecular dataset derived from MoleculeNet\cite{DBLP:journals/corr/WuRFGGPLP17}. It consists of molecular graphs where nodes represent atoms and edges signify chemical bonds. The primary objective is to ascertain whether a molecule is capable of inhibiting HIV replication. The dataset is organized based on two domain selections: scaffold and size. The first criterion, Bemis-Murcko scaffold\cite{Bemis1996ThePO}, refers to the two-dimensional structural foundation of a molecule. The second criterion pertains to the number of nodes in a molecular graph, a fundamental structural characteristic. 
        \item \textbf{GOODZINC} is a real-world molecular property regression dataset from the ZINC database\cite{DBLP:journals/corr/Gomez-Bombarelli16}. It comprises molecular graphs, each with a maximum of 38 heavy atoms. The primary task involved is predicting the constrained solubility\cite{DBLP:conf/icml/JinBJ18, DBLP:conf/icml/KusnerPH17} of the molecules.
        \item \textbf{GOODSST2} is a real-world natural language sentimental analysis dataset adapted from Yuan et al.\cite{DBLP:journals/pami/YuanYGJ23}. The dataset forms a binary classification task to predict the sentiment polarity of a sentence. The domains are split according to the sentence length.
\end{itemize}

\section{Training and test metric curves}
\begin{figure}[ht]
    \centering
    \begin{subfigure}[b]{0.49\linewidth}
        \includegraphics[width=\linewidth]{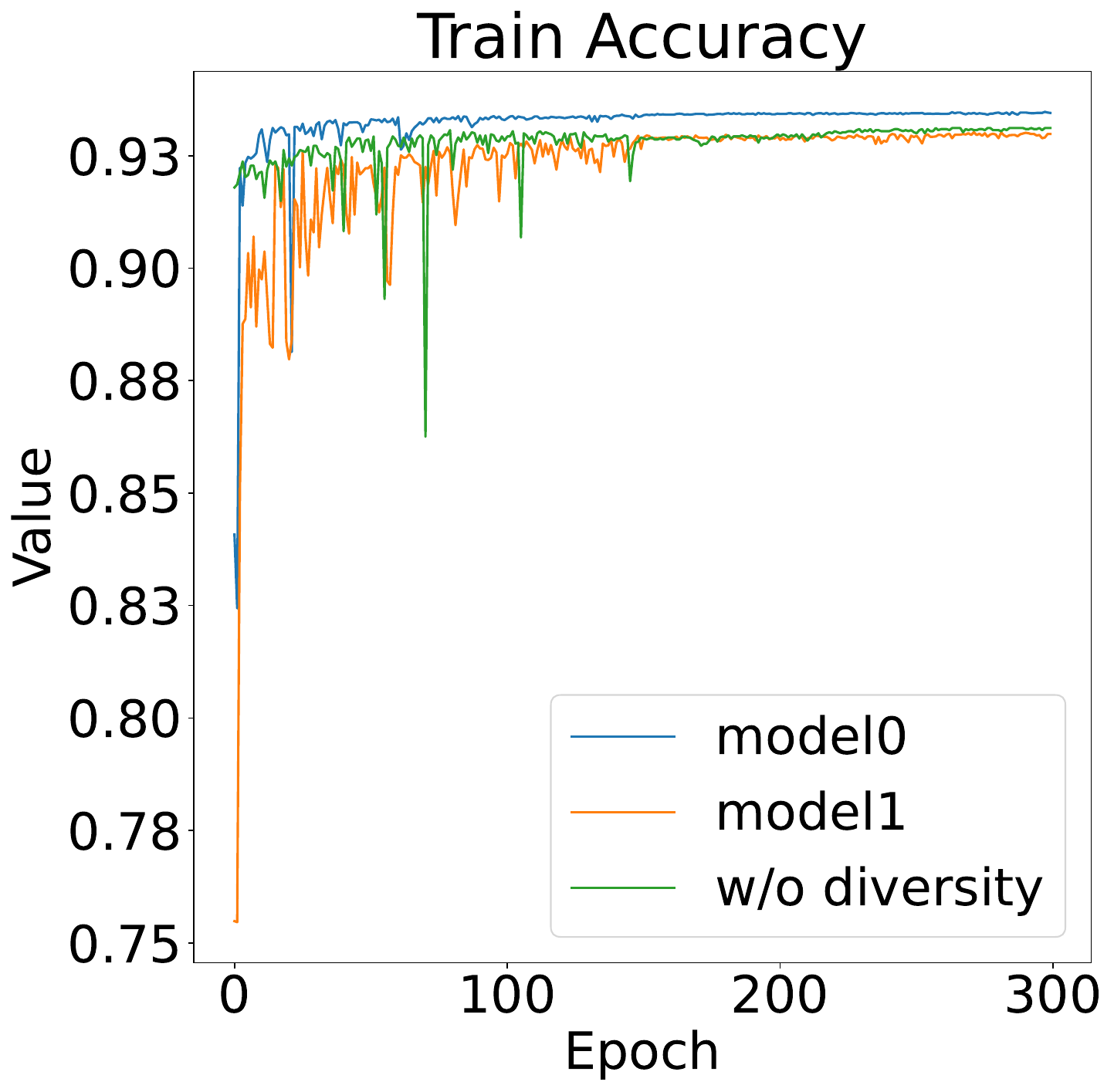}
        \caption{Train ACC on GOOD-Motif}
        \label{fig:sub1_}
    \end{subfigure}
    \hfill 
    \begin{subfigure}[b]{0.49\linewidth}
        \includegraphics[width=\linewidth]{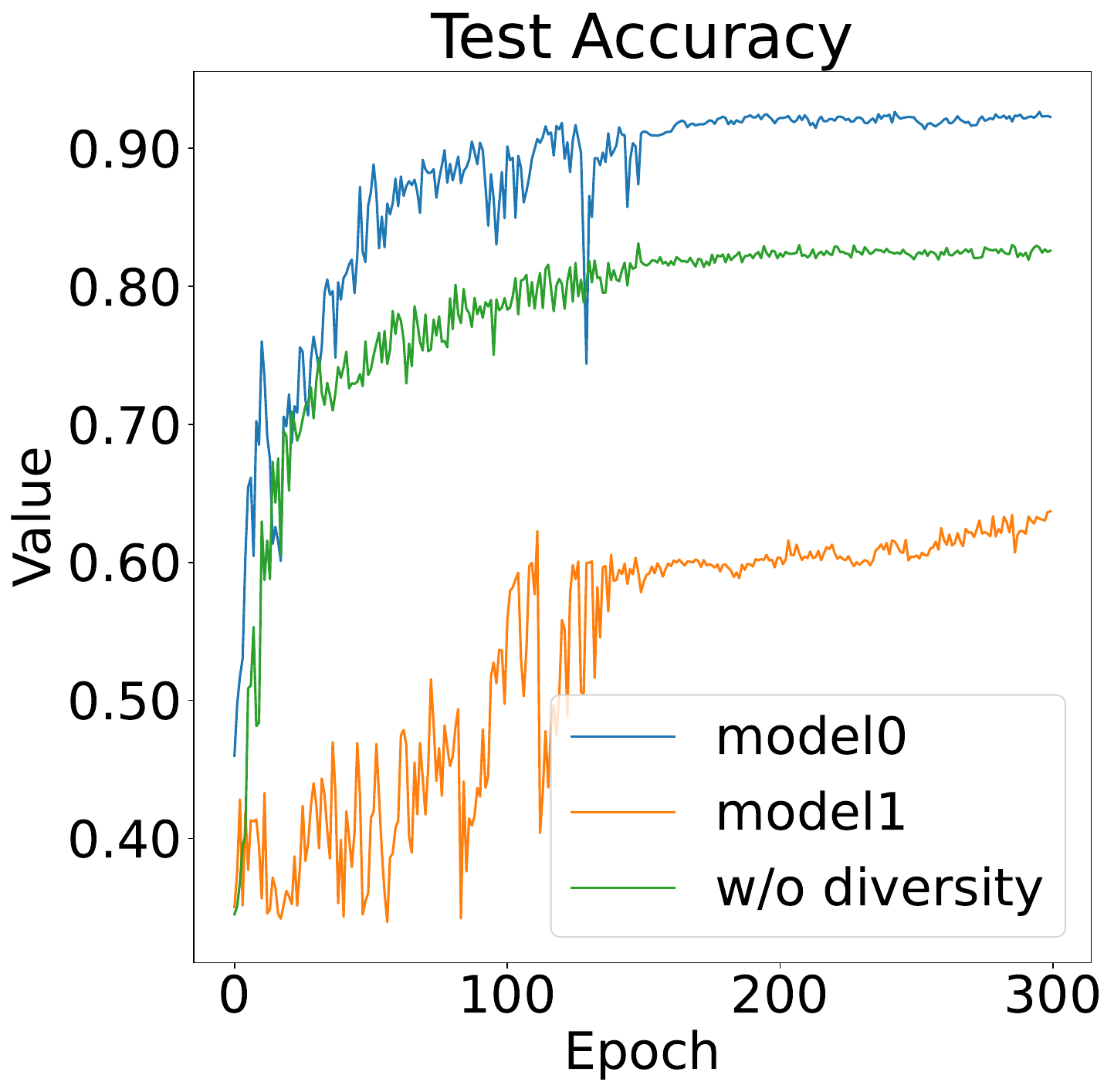}
        \caption{Test ACC on GOOD-Motif}
        \label{fig:sub2_}
    \end{subfigure}

    \begin{subfigure}[b]{0.49\linewidth}
        \includegraphics[width=\linewidth]{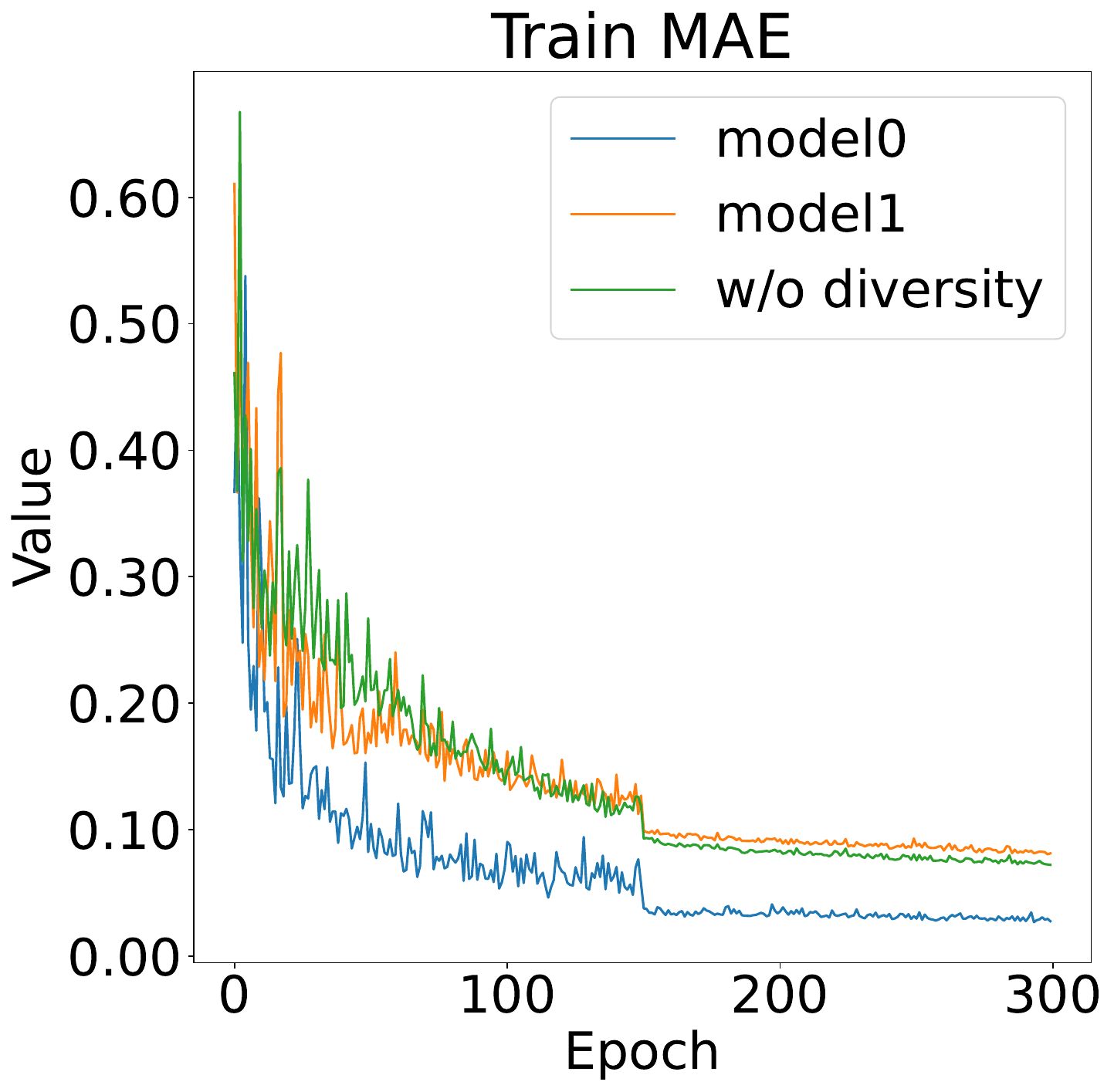}
        \caption{Train MAE on GOOD-ZINC}
        \label{fig:sub3_}
    \end{subfigure}
    \hfill
    \begin{subfigure}[b]{0.49\linewidth}
        \includegraphics[width=\linewidth]{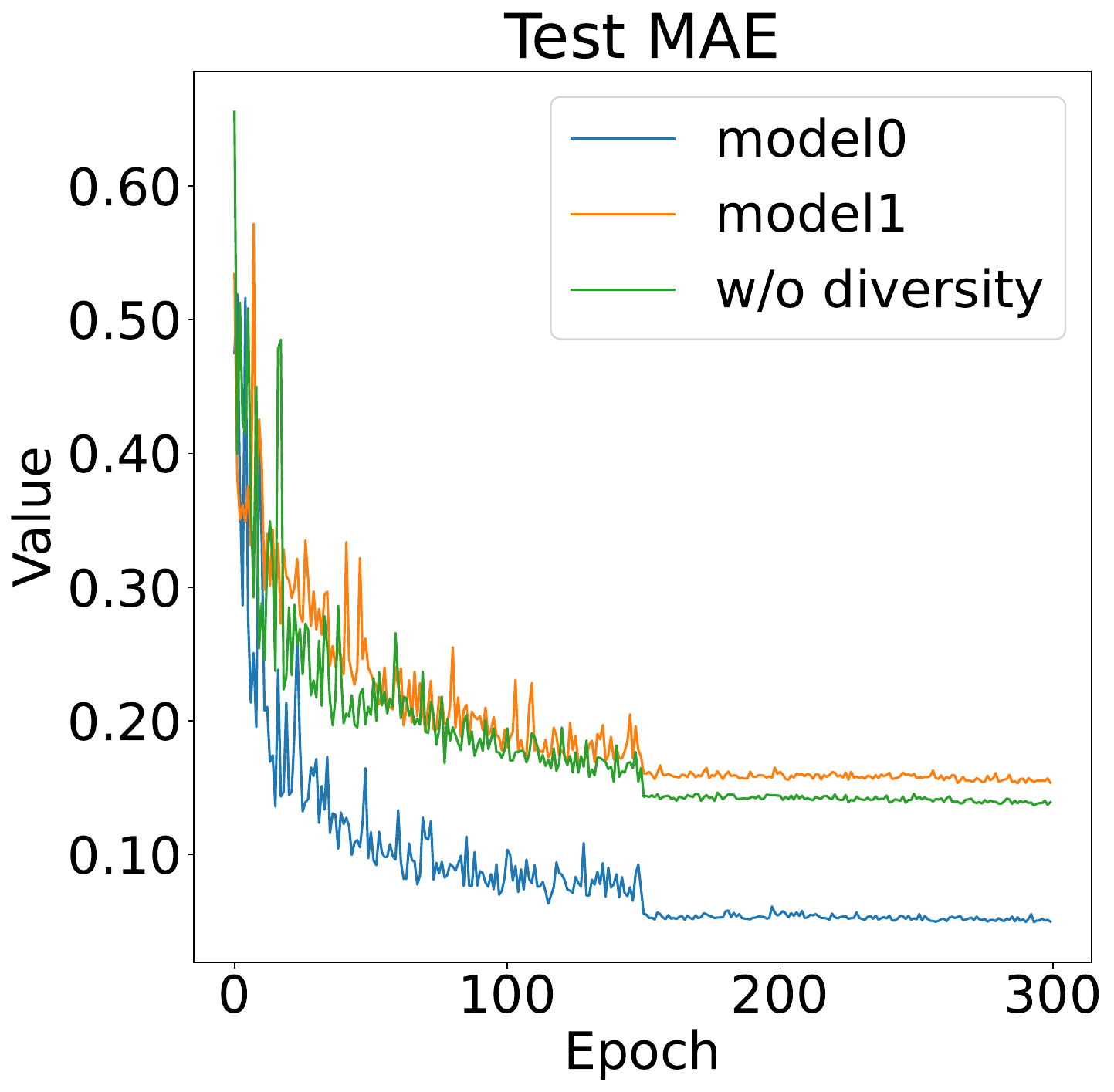}
        \caption{Test MAE on GOOD-ZINC}
        \label{fig:sub4_}
    \end{subfigure}        
    
    \caption{Train and Test metric curve of different models in the collections.}
    \label{fig:test}
\end{figure}

Figure~\ref{fig:test} shows the training and test metric curves on the GOODMotif (basis-concept scenario) and GOODZINC (scaffod-concept scenario). In the training stage, models in the collections can all achieve good performance and they attend to different label-predictive subgraphs. In the test stage, because the spurious correlation between the label and spurious subgraph diminish, only the model that attend to the real invariant subgraph can achieve a good performance.
\label{appendix: curve}

\section{Extra related work}
 \subsection{Graph long tail learning}
Graph long-tail learning\cite{chen2021topology, song2022tam, park2021graphens, yan2024rethinking} specifically deals with the problems posed by imbalanced data distributions where some classes of graph data are significantly underrepresented. This imbalance can mirror, and often exacerbates, the challenges faced in OOD generalization. Essentially, models trained on such skewed distributions might not only struggle with minority classes but also perform poorly when encountering unseen or novel distributions, as often happens in OOD scenarios.

\section{The results using ID validation set}
\label{appendix: idval}
Although the use of an out-of-distribution (OOD) validation set is a standard practice, and all baselines in our experiments adhere to this by employing the same OOD validation set, we have additionally conducted experiments using an in-distribution (ID) validation set to further demonstrate the effectiveness of DIVE. Our findings indicate that the improvements are even more pronounced when using the ID validation set. Specifically, the enhancement over CIGA is approximately 40\% in GOOD-Motif and around 10\% in GOODSST2. These results underscore the high precision of our current method when evaluated on the ID validation set. Our approach successfully extracts both spurious and invariant subgraphs, with the invariant subgraph predictor proving to be more indicative of both ID and OOD data (as depicted in Figure \ref{fig:test}, where the invariant predictor model 0 achieves superior performance on both the test and training sets). Consequently, even when using ID data for validation, a robust predictor that encapsulates invariant information is more readily identified, and the performance does not significantly deteriorate compared to using an OOD validation set.

\end{document}